\documentclass{article}

\usepackage{PRIMEarxiv}

\usepackage[utf8]{inputenc} 
\usepackage[T1]{fontenc}    
\usepackage{hyperref}       
\usepackage{url}            
\usepackage{booktabs}       
\usepackage{amsfonts}       
\usepackage{nicefrac}       
\usepackage{amsmath}
\usepackage{microtype}      
\usepackage{lipsum}
\usepackage{fancyhdr}       
\usepackage{graphicx}       
\graphicspath{{media/}}     
\usepackage{xcolor}
\usepackage{listings}
\usepackage{multirow}

\usepackage[round]{natbib}

\definecolor{codegreen}{rgb}{0,0.6,0}
\definecolor{codegray}{rgb}{0.5,0.5,0.5}
\definecolor{codepurple}{rgb}{0.58,0,0.82}
\definecolor{backcolour}{rgb}{0.95,0.95,0.92}

\lstdefinestyle{mystyle}{
    backgroundcolor=\color{backcolour},   
    commentstyle=\color{codegreen},
    keywordstyle=\color{magenta},
    numberstyle=\tiny\color{codegray},
    stringstyle=\color{codepurple},
    basicstyle=\ttfamily\footnotesize,
    breakatwhitespace=false,         
    breaklines=true,                 
    captionpos=b,                    
    keepspaces=true,                 
    numbers=left,                    
    numbersep=5pt,                  
    showspaces=false,                
    showstringspaces=false,
    showtabs=false,                  
    tabsize=2
}
\title{WERank: Towards Rank Degradation Prevention for Self-Supervised Learning Using Weight Regularization
}

\author{
Ali Saheb Pasand$^{*}$ \\
University of Waterloo
\And
Reza Moravej$^{*}$ \\
University of Toronto 
\And
Mahdi Biparva$^{\dagger}$ \\
Huawei Noah's Ark Lab
\And
Ali Ghodsi \\
University of Waterloo \\ \\
$*$ done during internship $\dagger$ correspondence to \texttt{mahdi.biparva@huawei.com}
}

\begin{document}

\maketitle

\begin{abstract}

A common phenomena confining the representation quality in Self-Supervised Learning (SSL) is dimensional collapse (also known as rank degeneration), where the learned representations are mapped to a low dimensional subspace of the representation space. 
The State-of-the-Art SSL methods have shown to suffer from dimensional collapse and fall behind maintaining full rank. 
Recent approaches to prevent this problem have proposed using contrastive losses, regularization techniques, or architectural tricks. 
We propose WERank, a new regularizer on the weight parameters of the network to prevent rank degeneration at different layers of the network.
We provide empirical evidence and mathematical justification to demonstrate the effectiveness of the proposed regularization method in preventing dimensional collapse. 
We verify the impact of WERank on graph SSL where dimensional collapse is more pronounced due to the lack of proper data augmentation. 
We empirically demonstrate that WERank is effective in helping BYOL to achieve higher rank during SSL pre-training and consequently downstream accuracy during evaluation probing.
Ablation studies and experimental analysis shed lights on the underlying factors behind the performance gains of the proposed approach.

\end{abstract}

\section{Introduction}

The goal of Self-Supervised Learning (SSL) is to learn useful representations of data without relying on human annotations. Recent advances in SSL have shown that it is possible to learn self-supervised representations that are competitive with supervised labels in a variety of settings including visual and graph domains ~\cite{NEURIPS2019_ddf35421,grill2020bootstrap,chen2020simsiam,bardes2022vicreg,zbontar2021barlow,ermolov2021whitening,DBLP:conf/icml/ChenK0H20}. 
SSL approaches enforce the encoding model to learn similar representations for semantically similar inputs. However, enforcing similarity between similar points alone may lead the model to trivially learn to output a similar embedding vector for every input. This phenomena, known as complete collapse, is undesirable since it provides no gradients for learning, and the representations offer no information for the downstream task.

Though complete collapse can be prevented in principled ways~\cite{DBLP:conf/icml/ChenK0H20,grill2020bootstrap,chen2020simsiam,bardes2022vicreg,zbontar2021barlow,ermolov2021whitening}, it is still common for the State-of-the-Art (SoTA) SSL methods to map representations to a low-dimensional subspace of the representation space. Avoiding this kind of dimensional (partial) collapse has remained a challenging problem across different SSL approaches \cite{NEURIPS2022_aa56c745,garrido2023rankme}. 
Dimensional collapse is linked with strong correlations between axes, which results in relatively uninformative embeddings \cite{DBLP:journals/corr/abs-2105-00470}. It has been shown that learning dimensionally collapsed, or rank-deficient, representations is a bottleneck for SSL models to achieve their best performance on downstream tasks \cite{NEURIPS2022_aa56c745,garrido2023rankme}.

Current SSL approaches aim to alleviate rank degradation in the final embedding space. However, they fail to address it particularly at early stages of the representation learning. In other words, enforcing a collapse prevention mechanism on the outputs of the final layer of deep networks does not necessarily prevent rank degradation at earlier layers \cite{bardes2022vicreg}. The implicit regularization present in deep networks causes dimensional collapse across layers of the network \cite{Jing2021UnderstandingDC}. Thus, the low rank solution found in an early layer would propagate to deeper layers. To address this problem, we propose \textbf{WERank}, a new \textbf{W}eight r\textbf{E}gularization term which serves as a complementary \textbf{Rank} degradation prevention mechanism on top of any SSL method. WERank prevents dimensional collapse throughout the network rather than the final layer. Unlike previous information maximization approaches \cite{bardes2022vicreg,zbontar2021barlow} employing variance/covariance/cross-covariance terms, WERank is directly computed on the weights of the neural network and is computationally more efficient.

We demonstrate the impact of WERank in the graph domain, where we show that WERank is generally effective in helping the model achieve higher downstream accuracy. We choose to study the impact of WERank on BGRL \cite{thakoor2022largescale} (the graph-based counterpart of BYOL \cite{grill2020bootstrap}). Though BGRL is among the best performing SSL methods, it is particularly prone to dimensional collapse since it does not employ an explicit mechanism to prevent rank degradation. \cite{tian2021understanding} argue that the dynamics of the alignment of eigenspaces between the predictor and its input correlation matrix play a role in preventing complete collapse. However, this mechanism does not explicitly force the model to prevent dimensional collapse. Additionally, we conduct experiments under the low augmentation regime, which is of particular importance in the graph domain. Previous work \cite{articleAFGRL,wang2023singlepass} have tried to minimized the reliance of graph SSL methods on augmentation. Unlike other modalities, where the augmentations carry meaningful semantic information, graph augmentations can be viewed as random noise, making graph SSL susceptible to rank degradation. We summarize our contributions as:

\begin{enumerate}
  \item We introduce WERank, a novel regularization term on the wights of the network to prevent dimensional collapse. We provide mathematical justification for the derivation of WERank.
  \item We demonstrate the impact of WERank on a toy dataset under two major situations which cause dimensional collapse; (i) namely when the model is overparameterized, and (ii) when the augmentation magnitude is strong. Our experiments demonstrate that WERank is effective in preventing dimensional collapse resulting from overparameterization. Besides, we show that the effectiveness of WERank increases as the augmentation magnitude is reduced.
  \item We demonstrate the effectiveness of WERank in the graph SSL domain on BGRL, where we show WERank is effective in preventing dimensional collapse and helping the model achieve higher downstream accuracy.
\end{enumerate}

\section{Related work}

\textbf{Self-Supervised Representation Learning: } Various SSL methods take different approaches to alleviate collapse. Contrastive methods prevent collapse by introducing negative samples. The InfoNCE loss minimizes the similarity of the embeddings of different samples while maximizing the similarity of different views of the same sample. The method is most often applied to Siamese architectures where the two branches have identical architectures and weights \cite{DBLP:conf/icml/ChenK0H20}. Previous empirical and theoretical evidence has shown that the repulsive effect of negative samples is not enough to eliminate dimensional collapse \cite{garrido2023rankme,Jing2021UnderstandingDC}. Unlike contrastive methods which view each sample as a separate class, clustering-based methods group samples into clusters based on some similarity measure \cite{10.5555/3495724.3496555,DBLP:conf/icml/HuangDGZ19,DBLP:conf/nips/CaronMMGBJ20}. Clustering approaches force the embeddings of different samples to belong to different clusters on the unit sphere.  Clustering approaches can generally be viewed as contrastive learning at the level of clusters rather than individual samples.

Inspired by knowledge distillation \cite{Hinton2015DistillingTK}, distillation methods such as BYOL and SimSiam \cite{grill2020bootstrap,chen2020simsiam} prevent collapse via architectural tricks. Distillation approaches propose an online network along with a target network, where the target network is updated with an exponential moving average of the online network weights. \cite{tian2021understanding} investigate how this mechanism works without using negative samples. While the dynamics of the alignment of eigenspaces between the predictor and its input correlation matrix plays a role in preventing collapse, distillation methods have no explicit mechanism to avert dimensional collapse and are thus prone to rank degradation.

Explicit regularization or information maximization methods \cite{bardes2022vicreg,zbontar2021barlow,ermolov2021whitening} take a principled approach to alleviate collapse via maximizing the information content of the embeddings. The aim of such methods is to produce embedding variables that are decorrelated from each other, thereby preventing an informational collapse in which the variables carry redundant information. For example, Barlow Twins \cite{zbontar2021barlow} computes the cross-correlation matrix computed between the outputs of the two identical networks along the batch dimension and forces it to be close to the identity matrix. VICReg \cite{bardes2022vicreg} adds a \textit{covariance} term to the loss to remove dimensional correlation with minimize the off-diagonal covariant terms over a batch as well as a \textit{variance} term to maximize the volume of projected samples in the embedding space. Analogous to other SSL methods, information maximization methods have shown to be subject to dimensional collapse \cite{garrido2023rankme}.

\textbf{Explicit Dimensional Collapse in SSL:} Recent work have shown that dimensional collapse is a practical bottleneck confining the performance across multiple SSL methods. \cite{garrido2023rankme} measure dimensional collapse by estimating the embeddings' rank. Their experimental results verify that $(i)$ having a high ranked embedding is tightly linked with the quality of the representation and serves as a necessity for achieving good downstream performance. $(ii)$ SoTA SSL methods fail to achieve close to full rank in the embedding space when trained on Image-Net. \cite{NEURIPS2022_aa56c745} shows that the optimal VICReg representation can be made full-rank by carefully selecting the loss hyperparameters. However, achieving full rank with VICReg is difficult in practice. In addition, having a full ranked output embedding may not necessarily translate to full ranked representations in earlier layers of the network.

To better understand the root cause of dimensional collapse, \cite{Jing2021UnderstandingDC} study the dynamics of the network weights in contrastive pre-training. They identify that contrastive methods are prone to dimensional collapse due to two mechanisms: $(i)$ strong augmentations cause the embedding space covariance matrix to become low-rank. $(ii)$ the over-parametrization of linear networks causes the network to find low-rank solutions. More generally, previous work has both theoretically and empirically established that over-parametrized neural networks tend to find flatter local minima and derive low-rank solutions \cite{DBLP:journals/corr/abs-1905-13655}. 

\section{WERank Regularization} 
\subsection{Notation}
We consider SSL pre-training in the graph domain, where a graph $\mathcal{G}$ is defined by a tuple $(V, A, X)$ where $V$ is the set of nodes in the graph such that $N=|V|$, $A \in \mathbb{R}^{N \times N}$ is the adjacency matrix representing the node connectivities in terms of the edges in the graphs, and $X \in \mathbb{R}^{N \times D}$ are the node features. We apply two stochastic graph augmentations, $\mathcal{T}_1$ and $\mathcal{T}_2$ to $\mathcal{G}$, obtaining two views: $\mathcal{G}_1 = (V_1, A_1, X_1)$ and $\mathcal{G}_2 = (V_2, A_2, X_2)$.
The SSL model consists of an encoder $f_\theta$ and a decoder/projector $g_\phi$. We define the representations $H \in \mathbb{R}^{N \times K}$ as $H = f(\mathcal{G})$ and the embeddings as $Z \in \mathbb{R}^{N \times M}$, where $Z = g(f(\mathcal{G}))$. Graph SSL models are pre-trained to learn useful node representations by enforcing the embeddings of the same nodes in $\mathcal{G}_1$ and $\mathcal{G}_2$ to be similar. In general, Graph SSL approaches minimize a loss function of the form: 
$$\mathcal{L(\theta, \phi)} = \mathbb{E}_{v_1 \sim V_1, v_2 \sim V_2} sim(z_1, z_2)$$ where $sim$ is a similarity function such as cosine or euclidean similarity. After pre-training is completed, the projector $g_\phi$ is often thrown away. The representations learned by the trained encoder $f_{\theta}$ can be used on any graph downstream task, such as node classification or edge prediction.

Dimensional collapse occurs when the embedding vectors span a low-dimensional subspace of $M$, and embedding dimensions have strong correlations and contain redundant information. In other words, the rank of $Z$ is reduced to be much smaller than the embedding dimension $M$. Contrastive loss functions prevent collapse by pushing dissimilar pairs apart \cite{DBLP:conf/icml/ChenK0H20} and distillation methods leave the loss unmodified but prevent collapse using architectural tricks \cite{grill2020bootstrap,chen2020simsiam}. On the other hand, information maximization methods prevent collapse by enforcing decorrelation between the axes of the embedding vectors in $Z$ via an explicit regularization term in the loss function \cite{bardes2022vicreg,zbontar2021barlow,ermolov2021whitening}. Covariance decorrelation makes all the components of embedding in $Z$ linearly independent from each other, encouraging different dimensions to represent different semantic content. However, the foundational issue with current feature decorrelation losses is that the regularization term is applied too late in the process. Thus, it is common to get low rank output embeddings even when the SSL method explicitly enforces feature decorrelation in the loss \cite{garrido2023rankme}. Furthermore, achieving full rank in earlier layers is of particular importance since the output representations from earlier layers (before the decoder) are used for the downstream task. In the following section, we derive WERank from the feature decorrelation losses in  methods.

\subsection{Rank Degradation Prevention by Weight Regularization}

Consider a neural network with $L$ trainable weight matrices $W_1, W_2, ..., W_L$. We can write the output of layer $l$ with input $X^{(l)}$ as:
$$X^{(l+1)} = \sigma(X^{(l)}W_l) = \sigma(H^{(l)}W_l).$$ 
For a Graph Convolution layer, the output can be written as: 
$$X^{(l+1)} = \sigma(D^{1/2}AD^{-1/2}X^{(l)}W_l) = \sigma(H^{(l)}W_l)$$ 
Where $A$ is the adjacency matrix of the graph $\mathcal{G}$, $D$ is the diagonal degree matrix and $\sigma$ is a non-linearity. For simplicity, we ignore the non-linearity in our computations. Given an embedding dimension $M$, and $N$ data points, we can defined the covariance matrix $C \in \mathbb{R}^{M \times M}$ as:
\begin{equation*}
C = \frac{1}{N} \sum_{i=1}^{N} (\boldsymbol{z}_i - \overline{\boldsymbol{z}}) (\boldsymbol{z}_i - \overline{\boldsymbol{z}})^T
\end{equation*}
where $\overline{\boldsymbol{z}} = \sum_{i=1}^{N} \boldsymbol{z}_i/N $. Similar to information maximization methods, dimensional collapse can be alleviated by enforcing feature decorrelation in the embedding space. A straight forward way to achieve feature decorrelation is by introducing the Forbenious norm between the covariance matrix of the embedding vector and the identity as a regularizer to the loss function:
\begin{equation*}
    ||C - I_{(M\times M)}||_F
\end{equation*}
However, our objective is to present a regularizer on the encoder layers rather than on the final embedding outputs of the SSL decoder. This is in contrast with VICReg and Barlow Twins where regularization is applied through the loss terms on the final embedding vectors. We compute the regularization term on the weights of the network $W_1, W_2, ..., W_L$ instead of the final output, achieving the same impact as applying the above regularizer at every layer.

Denote $\lambda_i$ the $i_{th}$ eigenvalue of covariance matrix $C$, and $W \in \mathbb{R}^{d_{in} \times d_{out}}$ the weights of a single-layer model, where without loss of generality the input dimension is larger than the output dimension. We note that we can write the Forbenious norm between the covariance matrix of output and identity as a function of the singular values of the covariance matrix:
\begin{align*}
     ||C - I_{(d_{out}\times d_{out})}||_F = \sqrt{tr\bigg((C - I) (C - I)^T\bigg)}=
     \sqrt{\sum_{i=1}^{d_{out}} (\lambda^2_i(C) - 2\lambda_i(C) + 1)} =
     \sqrt{\sum_{i=1}^{d_{out}} (\lambda_i(C) - 1)^2}
\end{align*}

The $i_{th}$ eigenvalue of the covariance matrix can be bounded as follows:
\begin{align*}
  \lambda_i(C) = \lambda_i(W^TH^THW) \leq 
  \lambda_i(W^TH^TH) \lambda_{d_{out}}(W) \leq 
  \lambda_i^2(H) \lambda_{d_{out}}^2(W)
\end{align*}

Where $\sigma_{d_{out}}(W)$ and $\sigma_{1}(W)$ are the largest and smallest non-zero singular values of $W$ respectively. Thus, from the above we get:

\begin{align*}
||C - I_{(d_{out}\times d_{out})}||_F = \sqrt{\sum_{i=1}^{d_{out}} (\lambda_i(C) - 1)^2} \leq
\sqrt{\sum_{i=1}^{d_{out}}(\lambda^2_i(H)\lambda^2_{d_{out}}(W) - 1)^2}
\end{align*}

$||C - I_{(d_{out}\times d_{out})}||_F$ is minimized if $W^TW$ is as close as possible to the diagonal matrix with its $i_{th}$ diagonal element $d_i = \frac{1}{\lambda_i^2(H)}$. A perfect whitening is only possible if all singular values of the matrix $H$ are the same and we regularize $W^TW$ to be as close as possible to a diagonal matrix with diagonal elements $\frac{1}{\sigma^2_i(H)}$. \\

Since the input $H$ to the network is not controllable, the only way for controlling the eigenvalues of covariance is regularizing the weights during training. However, even in the case of having the same singular values, calculating this regularization term requires performing Singular Value Decomposition (SVD) for each batch and each layer during all epochs which makes it computationally intractable. Additionally, if we perform the training in batch format, the optimization will likely be unstable since the singular values will be different for each batch. We resort to making the identity as the target matrix for $W^TW$. This regularization term makes all $\sigma^2_i(W)$'s as close as possible to 1, and as a result, will prevent the degradation of the rank of input matrix. We thus propose the following regularization term applied to every layer of the network:
\begin{align*}
    \label{dinsum}
    \mathcal{L}_{reg} = \sum_{l=1}^{L}\alpha_l \Big|\Big|W^T_lW_l-I\Big|\Big|_F
\end{align*}
where $\alpha_i$ controls the intensity of regularization for different layers. If at any layer $l$, the output dimension $d_{out}$ is larger than $d_{in}$, we can replace the regularization term with $\Big|\Big|W_lW^T_l-I,\Big|\Big|_F$. Notice the regularization term whitens the matrix $W^T_lW_l$ or $W_lW^T_l$ with the largest rank $max(d_{in}, d_{out})$. WERank can serve as a complimentary rank degeneration prevention mechanism on top of any SSL method by simply being added to the original SSL loss term as follows:
\begin{equation}
    \mathcal{L} = \mathcal{L_{SSL}} + \mathcal{L}_{reg}
\end{equation}
However, we follow the experimental setup of BGRL \cite{thakoor2022largescale} to show how effective WERank is on SSL methods such as BYOL \cite{grill2020bootstrap} where collapse is prevented solely through architectural tricks. We leave the experimental evaluation on other SSL approaches to future work.

\begin{figure*}[t]
  \centering
    \includegraphics[width=0.27\columnwidth]{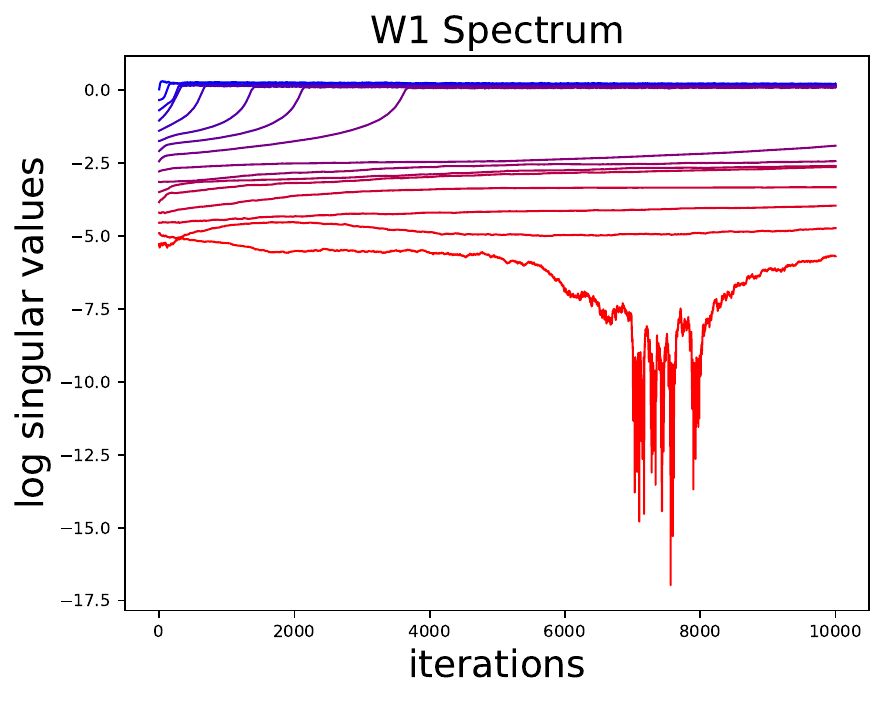}
    \includegraphics[width=0.27\columnwidth]{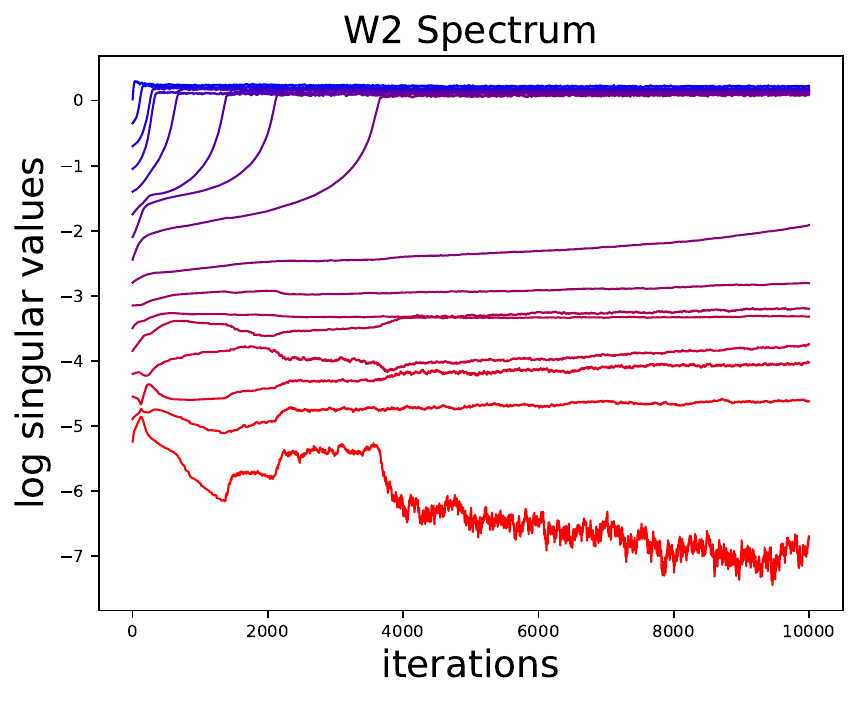}
    \includegraphics[width=0.27\columnwidth]{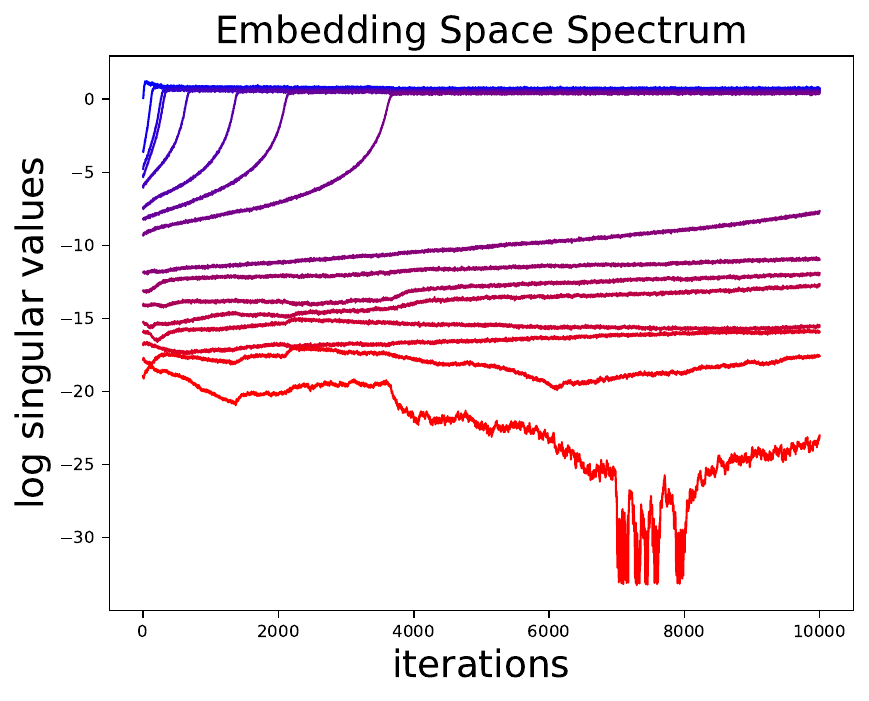} 
    \includegraphics[width=0.27\columnwidth]{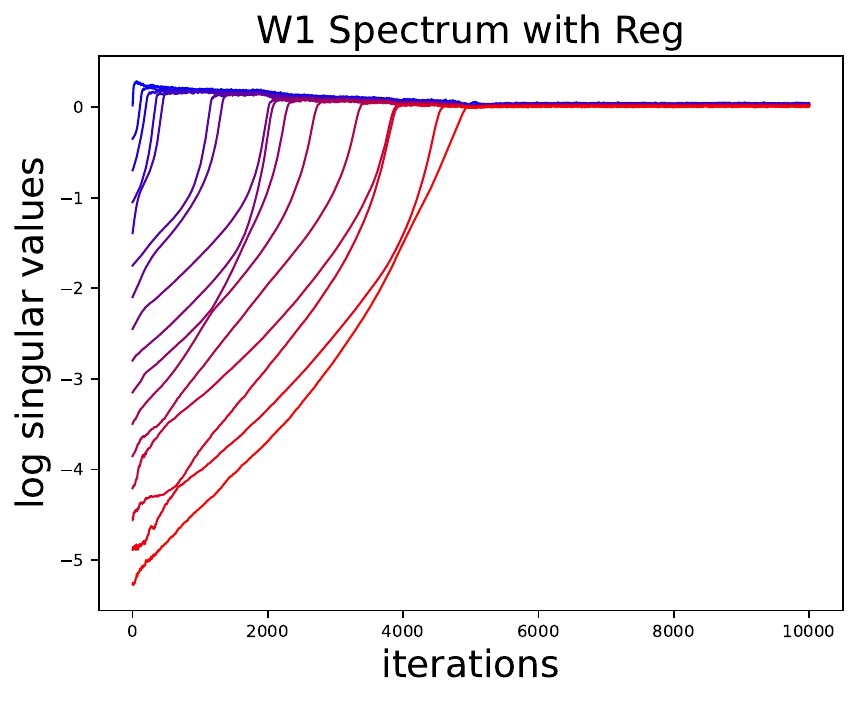}
    \includegraphics[width=0.27\columnwidth]{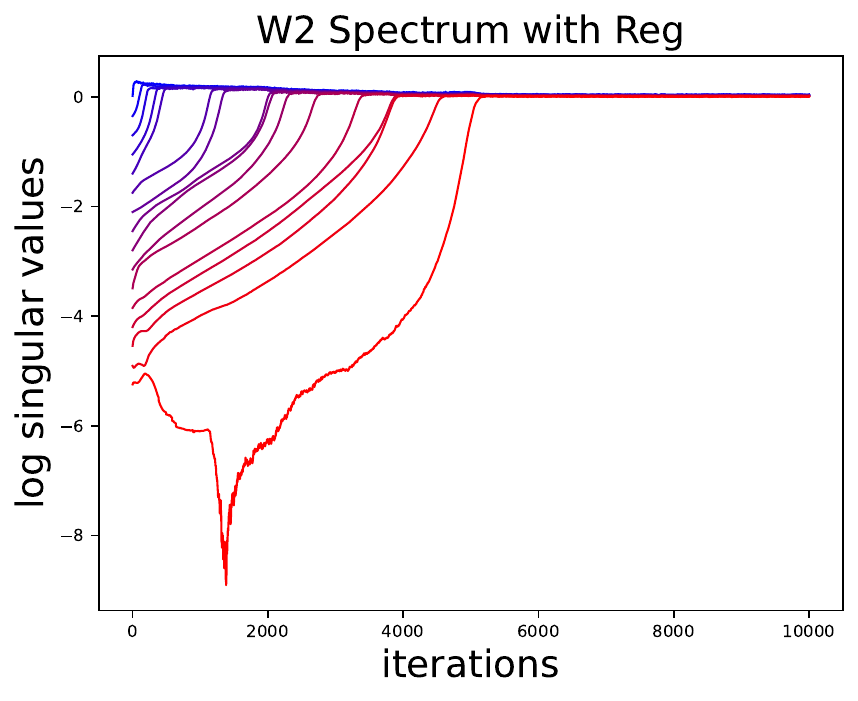}
    \includegraphics[width=0.27\columnwidth]{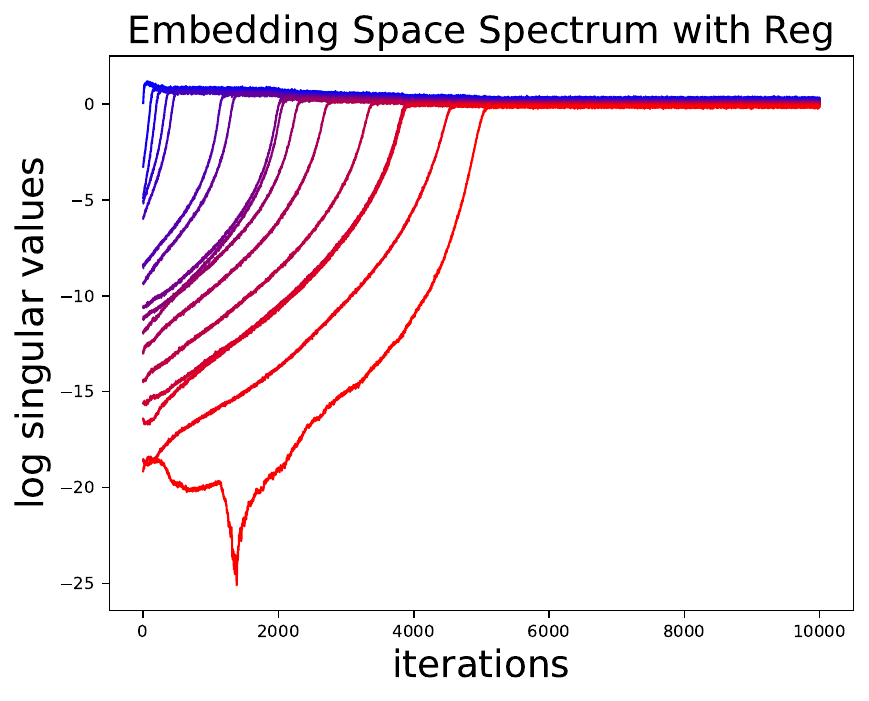}
  \caption{The singular values of the weight matrices and the embedding space covariance matrix during training (top) VICReg with no regularization (button) VICReg with the WERank regularizer. The augmentation magnitude ($k$) is set to $0.1$.}
  \label{fig1}
\end{figure*}

\section{Empirical Study on the Role of Weight Regularization}
To empirically support the effectiveness of WERank, we test the regularizer on SSL models under the two identified causes of collapse \cite{Jing2021UnderstandingDC}; namely $(i)$ implicit regularization caused by over-parameterization and $(ii)$ strong augmentation. Though \cite{Jing2021UnderstandingDC} identified the above causes of collapse for contrastive methods, we show that similar conditions can cause collapse in other SSL methods. It is possible to verify that the theoretical justification provided by \cite{Jing2021UnderstandingDC} can be extended to other SSL methods. 

Following the experimental setup by \cite{Jing2021UnderstandingDC}, we sample 1000 points from a 16 dimensional isotropic Gaussian with covariance matrix $\sum_{i,j} (x_i - x_j) (x_i - x_j)^T / N = I$. Two views of the sample $x_i$ are then generated from an additive Gaussian, with covariance matrix $\sum_{i} (x_i^{'} - x_i) (x_i^{'} - x_i)^T / N = \textit{block-diagonal}(0, k *I)$, where the block is $8 \times 8$. We consider two models with InfoNCE \cite{DBLP:conf/icml/ChenK0H20} and VICReg \cite{bardes2022vicreg} loss functions, as well as one model trained under the student-teacher Exponential Moving Average (EMA) algorithm (similar to BYOL \cite{grill2020bootstrap}). For each of the three models, we train an identical model with WERank added to the loss with a coefficient of $0.1$. Each model is trained for 1000 epochs in the full batch regime. We apply basic stochastic gradient descent without momentum or weight decay. Additional details about the hyperparameters are provided in Appendix. \\

The over-parameterization of neural networks can be a main cause for dimensional collapse. Due to the implicit regularization caused by over-parameterization, the smallest group of singular values grow significantly slower throughout training \cite{Jing2021UnderstandingDC}. First, we study the impact of WERank on the simplest over-parameterized setting by having a two-layer linear MLP with no bias. We denote the weight matrices of the network as $W_1, W_2 \in \mathbb{R}^{16 \times 16}$. Figure \ref{fig1} depicts the evolution of the singular values of the weight and embedding space covariance matrices of VICReg with and without WERank. The whitening loss applied by VICReg on the output of the final layer does not prevent dimensional collapse at earlier layers. However, WERank helps with preventing collapse resulted from the implicit regularization present in deep models by explicitly pushing the singular values up at every layer. We include similar results for InfoNCE and EMA models in Appendix. WERank is also effective in preventing rank degradation in deeper networks, where the impact caused by implicit regularization aggravates. Further results on model depth ablations can be found in Appendix.

\begin{figure}[t]
  \centering
    \includegraphics[width=0.32\columnwidth]{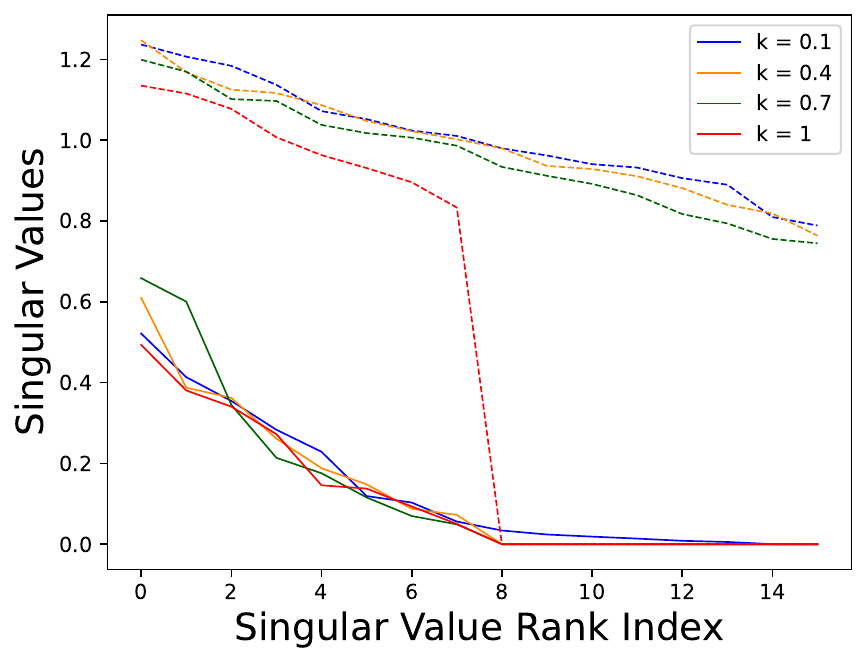}
    \includegraphics[width=0.32\columnwidth]{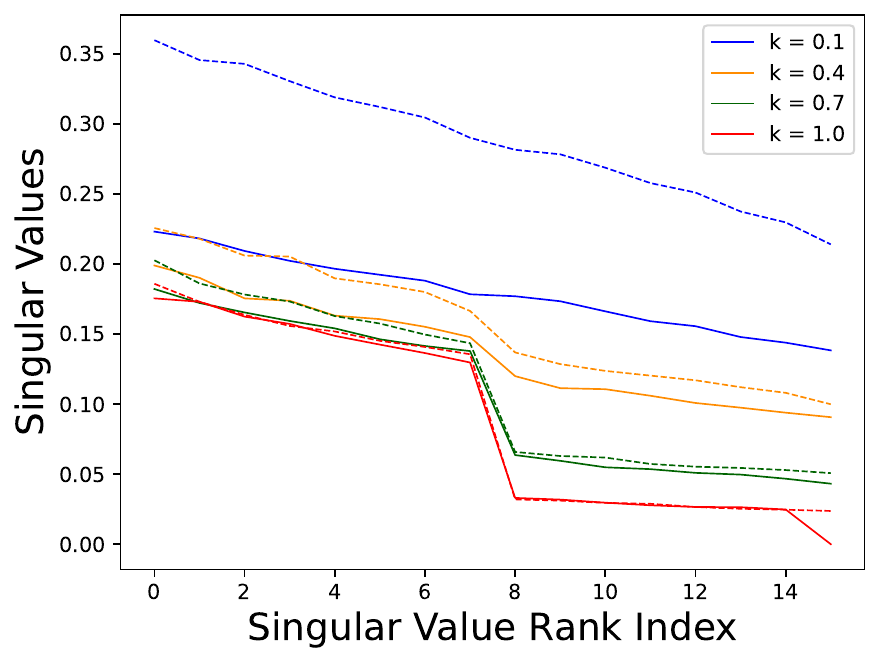}
    \includegraphics[width=0.32\columnwidth]{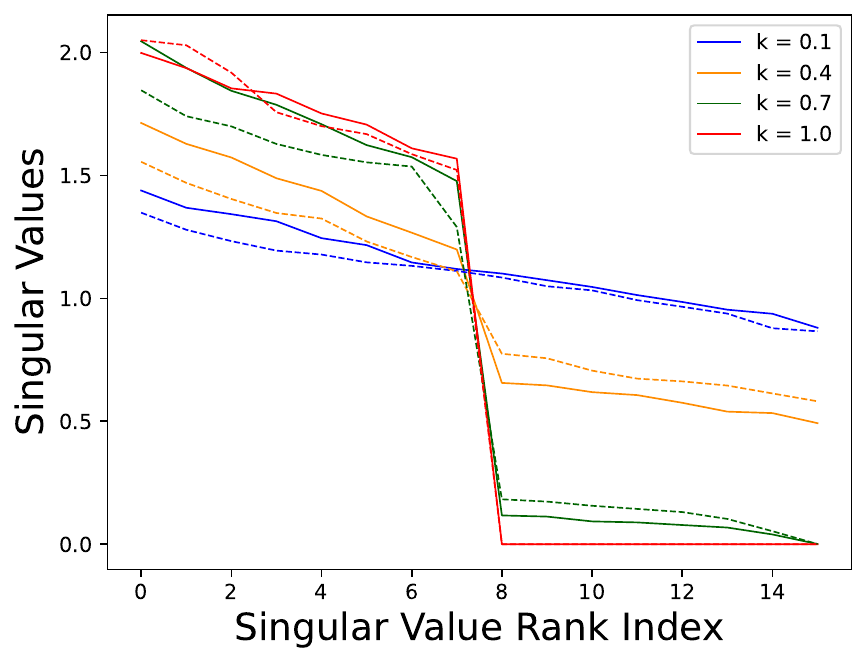}
  \caption{Weight matrix singular value spectrum with different augmentation amplitudes $k$, measured at the end of training. Solid lines depict the model with no regularizer and dotted lines depict model + WERank. (Left) EMA model (right) InfoNCE model (middle) VICReg model.}
  \label{fig2}
\end{figure}

In networks with limited capacity, strong augmentation along feature dimensions is a secondary cause for dimensional collapse \cite{Jing2021UnderstandingDC}. To test the impact of the magnitude of augmentation $k$, we choose a simple linear network with weights $W_1 \in \mathbb{R}^{16 \times 16}$. In Figure \ref{fig2}, we find the regularization term becomes less effective as the magnitude of augmentation increases \footnote{From an analytical lens, collapse in this case happens due to the dynamics of the time derivative of the weight matrix $W$ being determined by the augmentation distribution covariance matrix, refer to section 4.2 in \cite{Jing2021UnderstandingDC}}. This is expected, since extreme augmentation limits the amount of common information between the distorted views which can be used by the model for learning. Thus, the model will inevitably collapse in the presence of strong augmentation. However, WERank is distinctly impactful when the augmentation is weak. We note that the results for VICReg and VICReg + WERank aren't much different because WERank is simply enforcing the same variance/covariance terms as VICReg if only applied to a single layer. 

\section{Experimental Evaluation}

\begin{table*}[!t]
    \centering
    \resizebox{0.65\textwidth}{!}{
    \begin{tabular}{lcccccccc}
    \hline
        \textbf{Method/Dataset} & \textbf{Cora } & \textbf{CiteSeer } & \textbf{PubMed } & \textbf{DBLP }  \\ \hline
        \textbf{RandomInit-MLP} & 73.34 $\pm$ 3.29 & 62.82 $\pm$ 2.67 & 78.68 $\pm$ 1.06  & 76.54 $\pm$ 1.04   \\ 
        \hline
        \textbf{VICReg} & 88.56 $\pm$ 1.64 & 80.35 $\pm$ 1.63 & 86.86 $\pm$ 0.51 & 84.86 $\pm$ 0.51   \\ 
        \textbf{GBT} & 88.81 $\pm$ 1.32 & 81.13 $\pm$ 1.27 & 86.35 $\pm$ 0.59 & 85.55 $\pm$ 0.41  \\ 
        \textbf{GSwav} & 88.75 $\pm$ 1.31 & 79.41 $\pm$ 1.44 & 39.99 $\pm$ 0.67 & 81.19 $\pm$ 0.43  \\ 
        \textbf{Grace} & 89.54 $\pm$ 1.22 & 80.69 $\pm$ 1.27 & 39.99 $\pm$ 0.67 &  84.20 $\pm$ 0.10  \\ 
        \hline
        \textbf{BGRL} & 89.41 $\pm$ 1.24 & 81.47 $\pm$ 1.24 & 87.12 $\pm$ 0.45 & 85.57 $\pm$ 0.53   \\ 
        \textbf{BGRL+WERank} & 89.54 $\pm$ 1.29 & 82.44 $\pm$ 1.45 & 87.51 $\pm$ 0.52 & 85.73 $\pm$ 0.51 \\ 
        \hline
        \textbf{Supervised} & 81.13 $\pm$ 2.83 & 67.28 $\pm$ 3.34 & 82.43 $\pm$ 0.77 & 80.14 $\pm$ 1.02  \\ 
        \hline
    \end{tabular}
    }
    \label{dwnTable1}
    \caption{Classification accuracy along with standard deviations on small size datasets. \\
    }

\end{table*}

\begin{table*}[!t]
    \centering
    \small
    \caption{Classification accuracy along with standard deviations on medium size datasets. \\
    }
    \begin{tabular}{lcccccccc}
    \hline
        \textbf{Method/Dataset} & \textbf{Amz Comp} & \textbf{Amz Photos} & \textbf{CoCS} & \textbf{WikiCS} & \textbf{CoPhy}  \\ \hline
        \textbf{RandInit-MLP} & 84.56  $\pm$ 1.28  & 91.21  $\pm$ 1.22  & 91.05   $\pm$ 0.62  & 78.96  $\pm$ 0.38 & 91.64  $\pm$ 0.68  \\ 
        \hline
        \textbf{VICReg} & 91.59  $\pm$ 0.41  & 94.27  $\pm$ 0.65  & 93.42   $\pm$ 0.35  & 83.11  $\pm$ 0.24 & 95.61  $\pm$ 0.17  \\ 
        \textbf{GBT} & 90.83  $\pm$ 0.55  & 93.92  $\pm$ 0.62  & 93.49  $\pm$ 0.26 & 83.18  $\pm$ 0.19 & 95.46  $\pm$ 0.18  \\ 
        \textbf{GSwav} & 91.14   $\pm$ 0.518  & 94.47  $\pm$ 0.68  & 93.71  $\pm$ 0.28 & 83.65  $\pm$ 0.14 & 95.71  $\pm$ 0.15 \\ 
        \textbf{Grace} & 88.03   $\pm$ 0.71  & 94.57  $\pm$ 0.56  & 94.17  $\pm$ 0.31 & 83.41  $\pm$ 0.19 & OOM \\ 
        \hline
        \textbf{BGRL} & 91.79   $\pm$ 0.45  & 94.43  $\pm$ 0.58  & 93.77  $\pm$ 0.39 & 84.16  $\pm$ 0.12 & 95.67  $\pm$ 0.14  \\ 
        \textbf{BGRL+WERank} & 91.97   $\pm$0.51  & 94.54  $\pm$ 0.65  & 93.95 $\pm$ 0.42 & 84.18  $\pm$ 0.22  & 95.71 $\pm$ 0.14 \\ 
        \hline
        \textbf{Supervised} & 88.06  $\pm$ 1.28  & 91.68  $\pm$ 1.44  & 91.66   $\pm$ 0.77  &  76.54  $\pm$ 0.64 & 94.85  $\pm$ 0.34  \\
        \hline
    \end{tabular}
    \label{dwnTable2}
\end{table*}

We choose to to study the impact of WERank on BGRL which takes advantage of the student-teacher architecture. Unlike other SSL methods, BGRL has no explicit mechanism to prevent dimensional collapse other than some architectural tricks such as stop-gradient and exponential-moving-average encoders. We will empirically show evidence that such SSL methods can significantly benefit from explicit rank-degradation mechanisms such as WERank.

\subsection{Experimental Setup}
We follow the standard linear-evaluation protocol on graphs \cite{veličković2018deep,zhu2020deep,thakoor2022largescale}, where the graph encoder is trained in a fully unsupervised manner. A linear model is then trained on top of the frozen embeddings through a logistic regression loss with $l_2$ regularization, without back propagating on the weights of the encoder. Similar to prior work, we consider a GCN encoder \cite{kipf2017semisupervised} on the transductive tasks. We use MeanPooling encoders with residual connections for the inductive task of PPI. 

We perform our experiments on a variety graph datasets consisting of transductive and inductive tasks. We closely follow the same setup as \cite{thakoor2021bootstrapped} and \cite{bielak2022graph} on the following popular benchmark tasks: $(i)$ product category prediction on Amazon Computers and Photos $(ii)$ classifying Wikipedia articles (WikiCS) $(iii)$ classifying authors as nodes in citation networks (Coauthor CS/Physics) $(iv)$ classifying protein roles in protein-protein interaction (PPI) networks $(v)$ classification of subject areas of computer science articles on arXiv (ogb-arXiv) $(vi)$ classifying predicting article subject categories in citation networks including Cora, Citeseer, Pubmed, and DBLP. We include additional results including those on two hetreohpilic datasets in Appendix. Following prior work by \cite{zhu2020deep,thakoor2022largescale,veličković2018deep}, we generate distorted views of the input by dropping out edges and randomly masking node features. We closely follow the setup proposed by \cite{thakoor2022largescale} on homophilic datasets. For heterophilic datasets, we choose our own distortion factors by doing a simple grid search. Refer to Appendix for the augmentation factors used for each dataset and full dataset details and statistics.

\begin{figure}[!t]
  \centering
    \includegraphics[width=0.22\columnwidth]{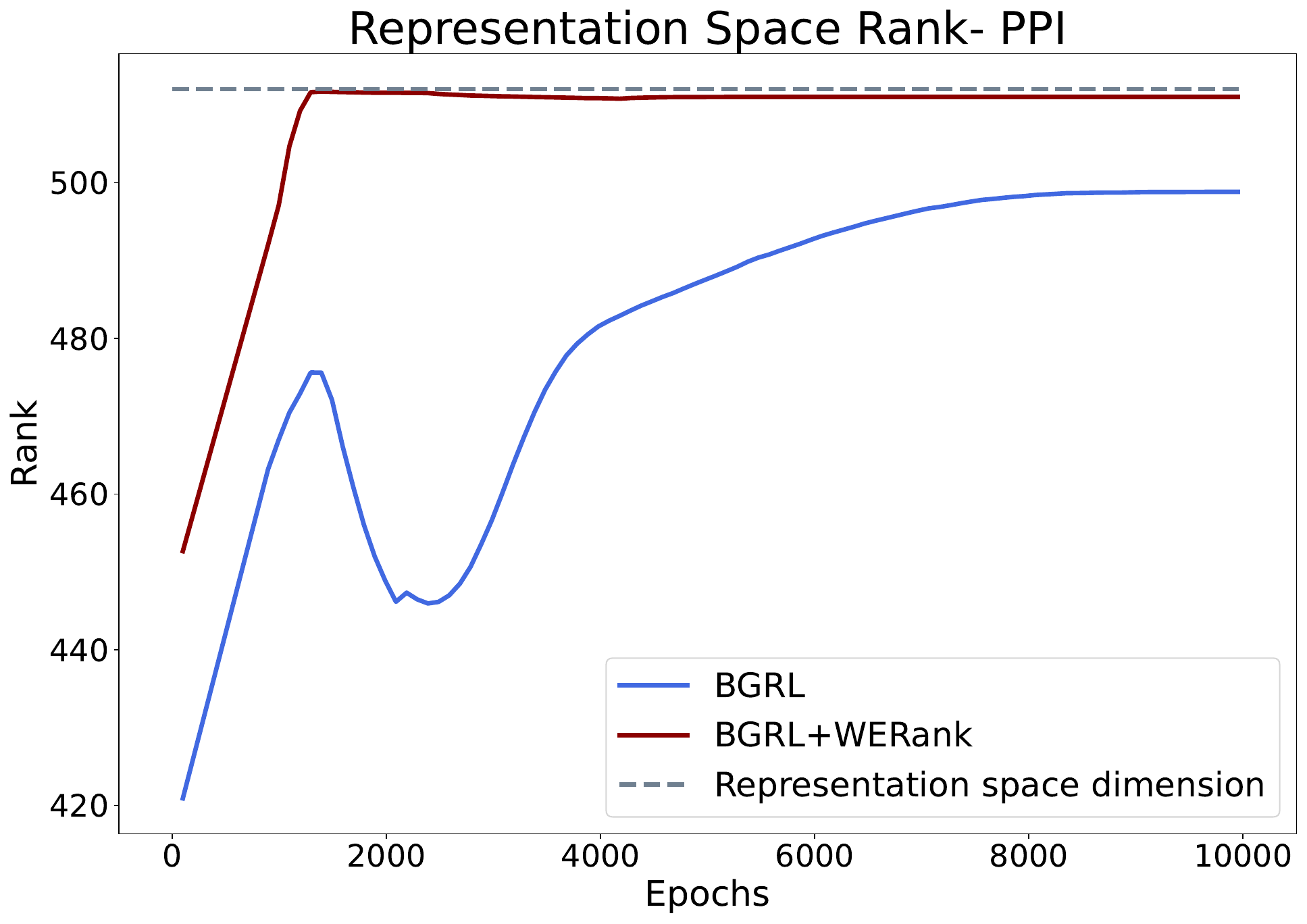}
    \includegraphics[width=0.22\columnwidth]{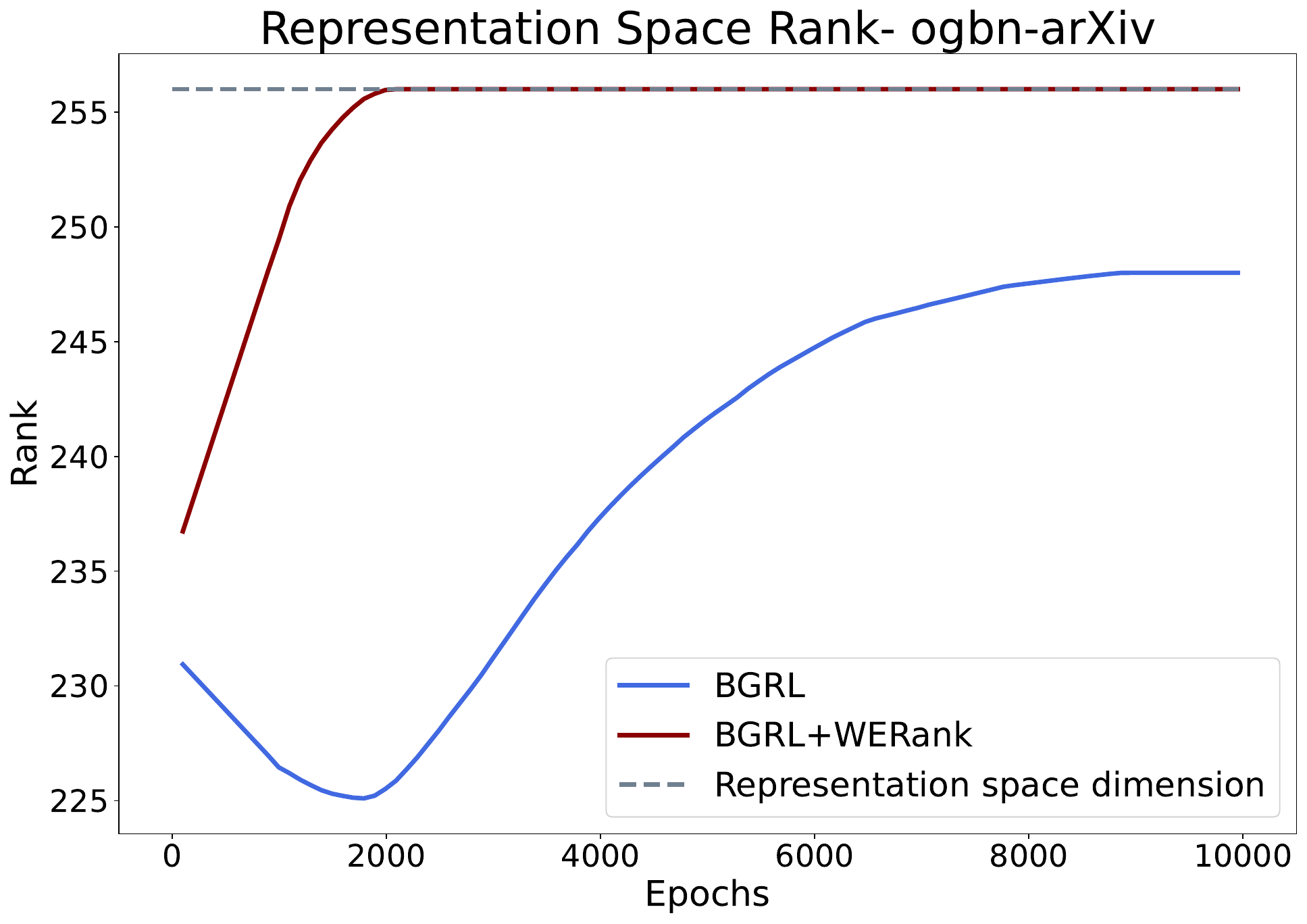}
    \includegraphics[width=0.22\columnwidth]{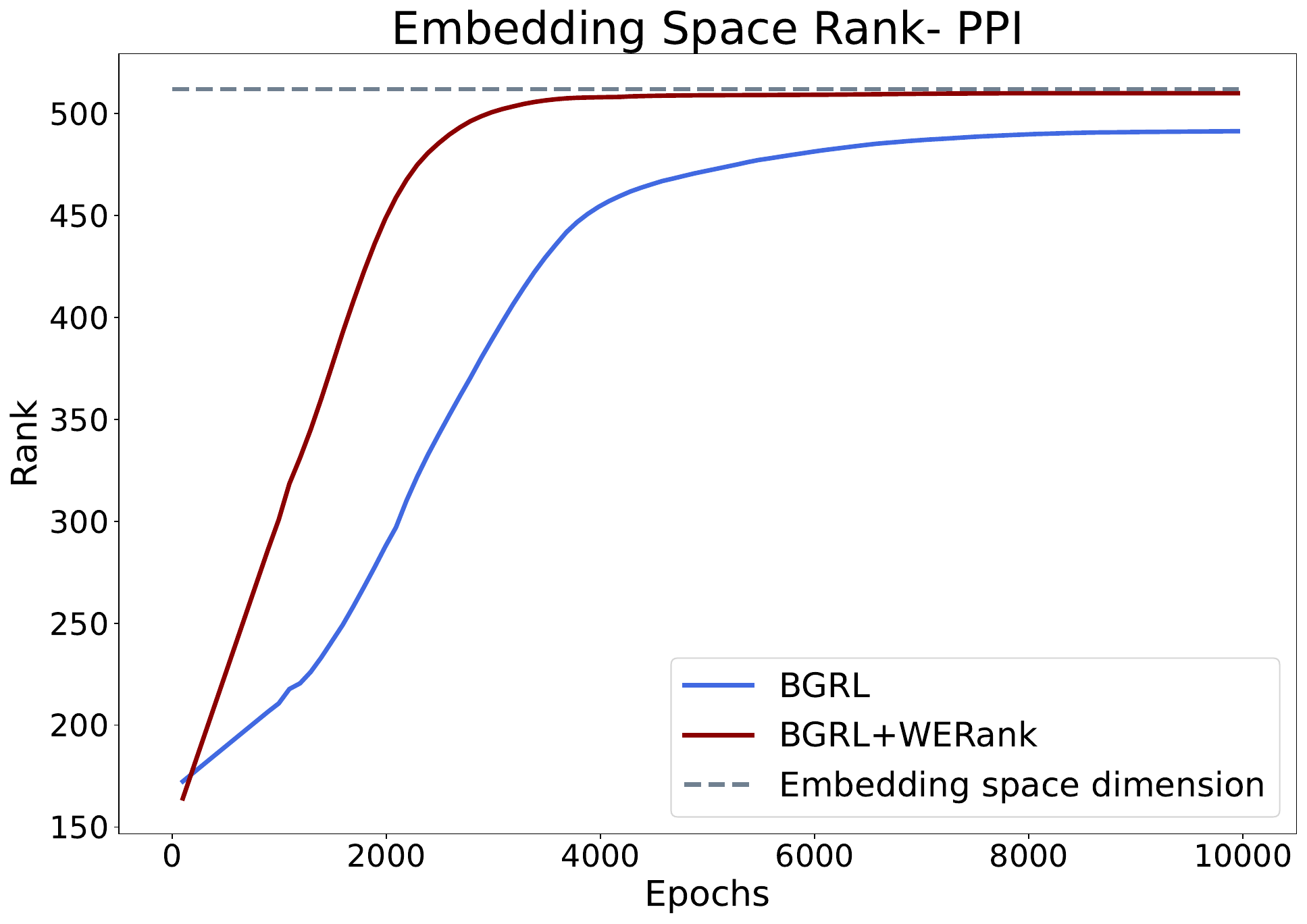}
    \includegraphics[width=0.22\columnwidth]{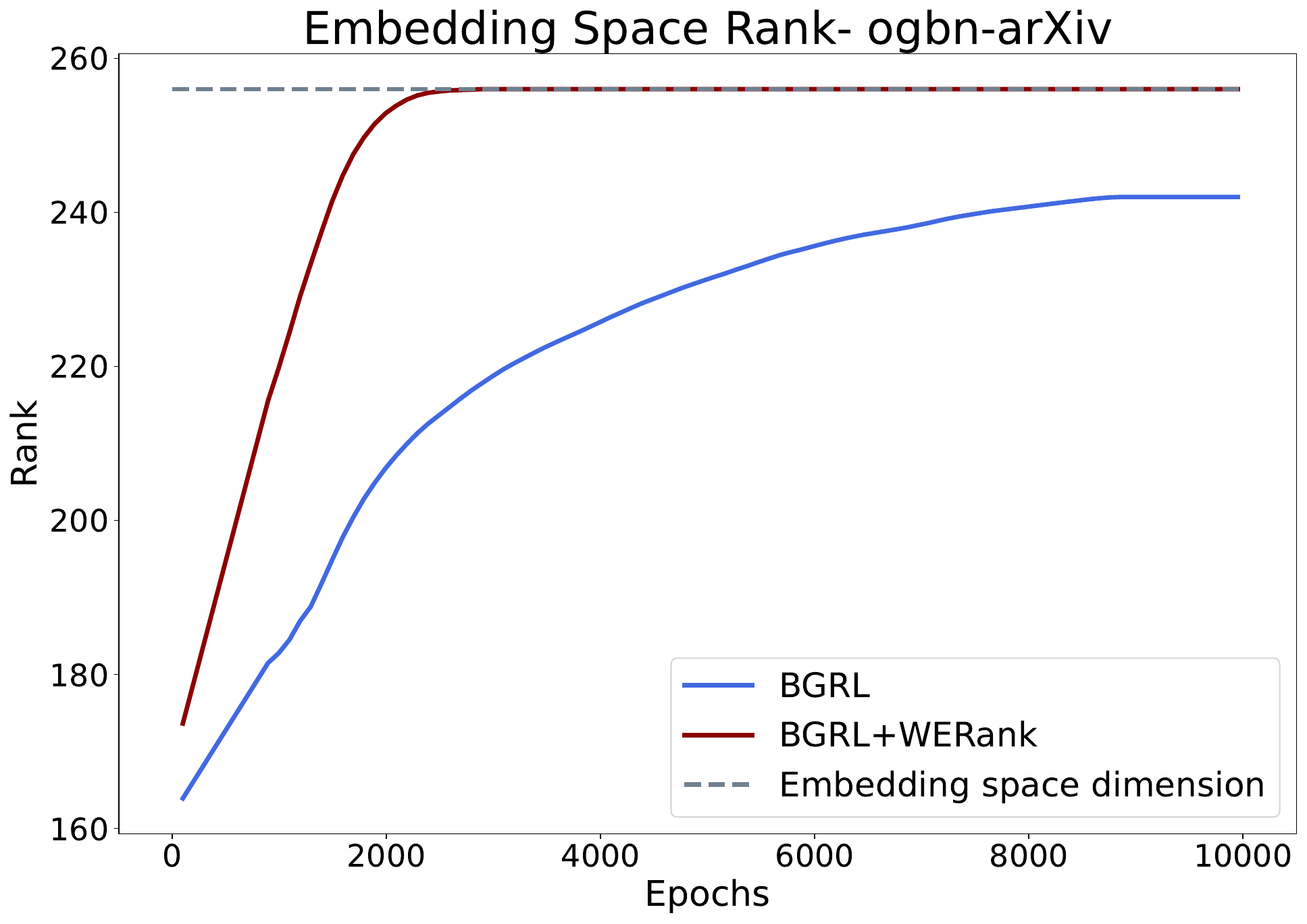}
\caption{The rank of the representation and embedding spaces on PPI and ogbn-arXiv. Curves over-smoothed for clarity.}
\label{rankPlot}
\end{figure}

We use popular graph SSL approaches as our baselines; Grace \cite{zhu2020deep}, BGRL \cite{thakoor2022largescale}, and Graph Barlow Twins \cite{Bielak_2022}. Each method is a modification to the original SSL method to suit the graph domain (SimCLR, BYOL and Barlow Twins respectively). We also adapt Swav \cite{caron2021unsupervised} and VICReg \cite{bardes2022vicreg} to the graph domain and use them as baselines. Additionally, we include results for a supervised GCN model and Random-Init which uses embeddings from a randomly initialized encoder. Random-Init results are indicative of the quality of the inductive biases present in the encoder model. 

\begin{table}[!t]
    \small
    \centering
    \caption{Classification accuracy along with standard deviations on the inductive task of PPI and large transductive task of ogbn-Arxiv . Our experiments are over 20 random dataset splits and model initializations.\\
    }
    \begin{tabular}{lcccccccc}
    \hline
        \textbf{Method/Dataset} & \textbf{PPI} &  \textbf{ogbn-arXiv}  \\ \hline
        \textbf{RandInit-MLP} & 77.73 $\pm$ 2.49 & 60.85 $\pm$ 0.79   \\ 
        \hline
        \textbf{VICReg} & 66.38 $\pm$ 0.02 &  61.28 $\pm$ 0.12 \\ 
        \textbf{GBT} & 78.10 $\pm$ 0.49 &  57.19$\pm$ 0.09  \\ 
        \textbf{GSwav} & 80.99 $\pm$ 0.03 &  71.36 $\pm$ 0.02 \\ 
        \hline
        \textbf{BGRL} & 68.98$\pm$ 0.02 &  71.74$\pm$ 0.03 \\ 
        \textbf{BGRL+WERank} & 69.07 $\pm$ 0.02 &  71.62 $\pm$ 0.05  \\ 
        \hline
        \textbf{Supervised} & 78.32 $\pm$ 2.01 &  71.53 $\pm$ 0.45  \\ 
        \hline
    \end{tabular}
    \label{dwnTable3}
\end{table}

For small and Amazon datasets, we don't perform any hyperparamter search for the WERank regularization coefficient and set the coefficient to $0.1$. For other medium and large size datasets, we perform a hyperparamter search over 7 different values (as depicted in Figure \ref{figAb1}). We apply the WERank regularization on top of the GCN and linear layers in the encoder, but don't apply it on top of the predictor. 
Our experimental results are averaged over 20 random dataset splits and model initialization seeds. OOM indicates running out of memory on a 16GB V100 GPU.

\begin{table*}[t]
  \centering
  \small
  \caption{Downstream Performance of BGRL and BGRL + WERank under different magnitudes of augmentation. Our experiments are over 20 random dataset splits and model initializations.\\}
  \resizebox{0.85\textwidth}{!}{%
\begin{tabular}{l|ccccccc} 
\hline
Dataset                           & Model         & 0.01 & 0.05 & 0.1 & 0.5 & 2  \\ 
\hline
\multirow{2}{*}{WikiCS}           & BGRL          &   $83.43 \pm 0.22$   &   $83.84 \pm 0.23$   &  $83.68 \pm 0.22$   &   $83.71 \pm 0.18$ &  $83.71 \pm 0.18$ \\
                                  & BGRL + WERank &   $\mathbf{83.96 \pm 0.15}$ & $\mathbf{84.01 \pm 0.19}$ &  $\mathbf{83.98 \pm 0.16}$  & $\mathbf{83.85 \pm 0.22}$     &  $\mathbf{83.84 \pm 0.20}$    \\ 
\hline
\multirow{2}{*}{CoPhy} & BGRL     &   $95.17 \pm 0.16$  &  $95.34 \pm 0.18$    &  $95.45 \pm 0.17$   &  $95.76 \pm 0.12$  &    $95.68 \pm 0.14$  \\
                                  & BGRL + WERank &   $\mathbf{95.22 \pm 0.12}$   &   $\mathbf{95.36 \pm 0.17}$   &  $\mathbf{95.50 \pm 0.19}$   &  $\mathbf{95.78 \pm 0.14}$   &   $\mathbf{95.70 \pm 0.14}$  \\ 
\hline
\multirow{2}{*}{CoCS}      & BGRL   &   $93.92 \pm 0.40$   &   $93.97 \pm 0.42$   &  $\mathbf{94.05 \pm 0.39}$   &   $\mathbf{93.90 \pm 0.40}$  &  $\mathbf{93.68 \pm 0.46}$ \\
                            & BGRL + WERank &   $\mathbf{93.99 \pm 0.41}$    &   $\mathbf{94.03 \pm 0.44}$   &   $93.40 \pm 0.41$  &   $93.81 \pm 0.31$  &   $93.52 \pm 0.34$   \\ 
\hline
\multirow{2}{*}{ogbn-arXiv}       & BGRL   &  $66.82 \pm 0.03$    &   $66.94 \pm 0.03$   &   $68.20 \pm 0.04$  &  $\mathbf{71.64 \pm 0.03}$   &  $68.11 \pm 0.03$  \\
                                  & BGRL + WERank &   $\mathbf{68.64 \pm 0.04}$   &  $\mathbf{70.03 \pm 0.02}$    &  $\mathbf{69.82 \pm 0.02}$   &  $71.43 \pm 0.02$ &   $\mathbf{68.60 \pm 0.03}$ \\
\hline
\end{tabular}
}
\label{augTable}
\end{table*}

\subsection{Experimental Results}
Experimental results indicate that WERank is generally effective in helping the model achieve higher downstream accuracy, though the downstream accuracy in the graph domain are not substantially different due to high inductive bias. WERank is effective in the embeddings carrying more information, which could be advantageous for the downstream task. 
First, Figure \ref{rankPlot} illustrates that WERank is effective in preventing rank degradation in BGRL. On both the representation and embedding spaces, the baseline SSL model cannot maintain a full-rank mapping over the course of pre-training while WERank is very effective to make the mapping full-rank. This illustrates the effectiveness of an explicit rank-degradation mechanism throughout various layers of representation learning.

Quantitative experimental evaluation on datasets with various scales reveals that WERank can improve the results of the baseline method, though the improvement is marginal. The experimental results are presented in Table \ref{dwnTable1}, \ref{dwnTable2}, \ref{dwnTable3} for small-, medium-, and large-scale datasets. Despite the marginal improvement, the consistency to improve is indicates that WERank is effective to impact the downstream task. However, improving the rank of the embedding space is a necessary but not sufficient condition to improve the performance of the downstream task \cite{garrido2023rankme}. We leave the extension of WERank to other data modalities such as images to future work.

\subsection{Ablations Studies and Empirical Analysis}

\begin{figure}[t]
  \centering
    \includegraphics[width=0.5\textwidth]{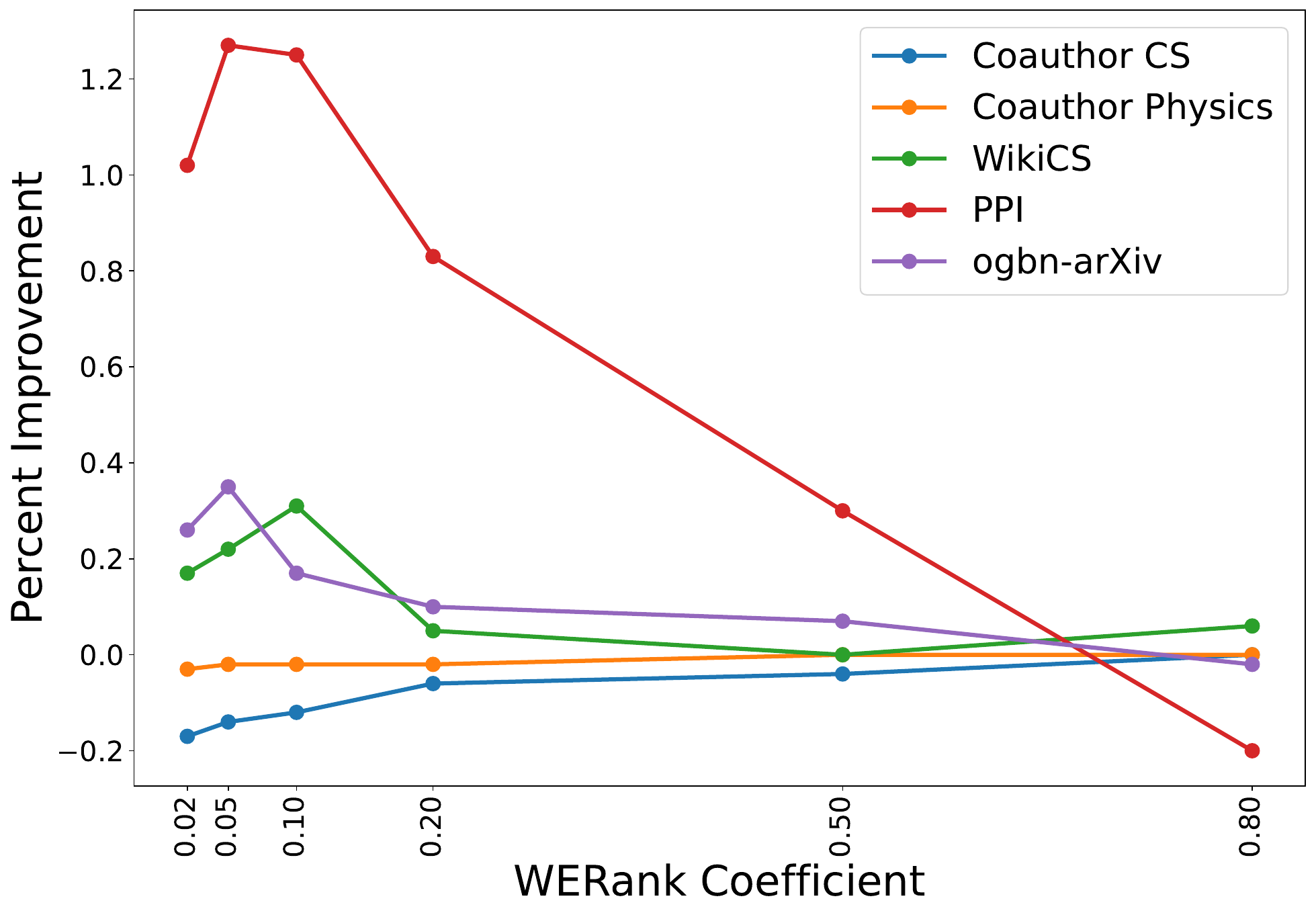}
  \caption{Percent improvement of BGRL + WERank under different coefficients over BGRL + WERank with coefficient 1. The same coefficient $\alpha$ is applied to every layer in the encoder.}
  \label{figAb1}
\end{figure}

We train BGRL+WERank with various values for $\alpha$  in \{0.02, 0.05, 0.1, 0.2, 0.5, 0.8, 1\} to test the impact of the WERank regularization coefficient $\alpha$ on the performance of the model. Figure \ref{figAb1} shows the percent improvement of each model against BGRL+WERank with $\alpha = 1$ (exact numbers can be found in Appendix). We find that a higher emphasis on the WERank regularizer does not translate to better performance on PPI and ogbn-arXiv. Since the node feature dimensions in PPI and ogbn-arXiv are smalll (50 and 128 respectively), a low dimensional (collapsed) representation space would be enough to embed key information. Thus, BGRL with no regularizer is enough to learn informative representations on these datasets. On the other hand, increasing the regularizer coefficient results in better performance on Coauthor CS and Coauthor Physics, where the feature dimensions are high (6,805 and 8,415 respectively). The relationship between node features and the impact of WERank could explain the result in Table \ref{dwnTable3}, where BGRL with WERank has a hard time improving the baseline BGRL on datasets with small number of node features. Additionally, we speculate that higher number of samples in a graph would help the model learn more informative representations. Thus, BGRL alone seems to be sufficient on larger graphs, where WERank is more impactful on graphs with fewer nodes.

To further investigate the impact of augmentation strength on WERank, we test the performance of Graph SSL models with and without WERank under different magnitudes of augmentation. Denote $p_{f,i}$ and $p_{e,i}$ the node feature masking and edge dropout augmentation factors of a dataset. For each dataset, we multiply $p_{f,i}$ and $p_{e,i}$ by one of $0.01, 0.05, 0.1, 0.5, 1, 2$. For example, $p_{f,1}$, and $p_{e,1}$ are set to $0.3$, and $0.4$ for the Coauhtor CS dataset. Hence, we would train 12 models under 6 different augmentation settings; BGRL with and without WERank trained on virtually no augmentation ($p_{f,1}= 0.3 \times 0.01$ and $p_{e,1}= 0.4 \times 0.01$) to very strong augmentation ($p_{f,1}= 0.3 \times 2$ and $p_{e,1}= 0.4 \times 2$). When the original augmentation parameter is larger than 0.5, we cap the augmentation factor to 1. We fix the WERank coefficient to 0.1 for all datasets. Similar to our ablations on the toy datasets (Figure \ref{fig2}), we find that WERank is most effective when the augmentation is weak. BGRL with WERank constantly outperforms the vanilla BGRL under the low distortion setting, where as the gap between the two models closes as the augmentation magnitude increases.

\section{Conclusion}
In this work, we proposed WERank to prevent rank degradation in self-supervised pre-training. It introduces a regularization term on the weights of the neural network and is added to the primary SSL pre-training loss function. WERank regulates the pre-training representation learning to prevent dimensional collapse in the hope of achieving more reliable representations for the downstream tasks. We provided empirical evidence showing that applying feature decorrelation via WERank in earlier layers of the network prevents rank degradation arising from implicit regularization in deep networks. Our results show that WERank is less effective when collapse happens as a result of extreme augmentation, and has a higher impact under weak augmentation. Our results on the graph domain show that WERank is generally effective in helping BGRL achieve higher accuracy, particularly when the augmentation is of low magnitude. We aim to study the impact of WERank on other domains in the future work.

\clearpage

\bibliography{arxiv}

\clearpage
\appendix

\section{WERank Implementation}
We include a simple PyTorch implementation for WERank. The parameter \texttt{loss} is the SSL loss function criterion (such as VICReg or InfoNCE loss). \texttt{network} is a PyTorch model, and \texttt{Z1},\texttt{Z2} are the embedding tensors obtained from passing the corresponding augmented views \texttt{X1}, \texttt{X2} to \texttt{network}. The final parameter \texttt{alphas} is regularization coefficients (list with length equal to the number of layers of the \texttt{network}). Our code implementation includes a normalization by a factor or $1/d^2$ at each layer.

\begin{lstlisting}[language=Python]
import torch

def ForbNorm(A):
    (torch.abs(covar - torch.eye(covar.shape[0]))).sum()

def WERank_loss(loss, network, Z1, Z2, alphas):
    Reg = 0
    for W, alpha in zip(net.parameters(), alphas):
        d = max(W.size())
        if W.shape[0] < W.shape[1]:
            cov = W @ W.T
        else:
            cov = W.T @ W
        F_Norm = ForbNorm(cov - torch.eye(d)) / (d**2)
        Reg += alpha * F_Norm
    return loss + Reg
\end{lstlisting}

\section{The Impact of WERank on the Learned Representations}

\subsection{WERank Encourages Orthonormality}
We provide a simple justification for why the the layer-wise regularization by WERank encourages an orthonormal representation space per layer. 

\textbf{\textit{Proposition: }} For a single layer of the neural network with weights $W \in \mathbb{R}^{d_{in} \times d_{out}}$, minimizing the regularization term $\Big|\Big|W^TW- I \Big|\Big|_F$ results in $\lambda_i(W)$ getting close to one for all $i=1,...,d_{out}$.

\textit{Proof:} Denote $U_{(W^TW)} \Lambda U^T_{(W^TW)}$ as the eigenvalue decomposition of $W^TW$ with $\lambda_i$ being the $i_{th}$ eigenvalue. We proceed by writing the regularization term in terms of the eigenvalues of $W^TW$:

\begin{equation}\label{eq:eq1}
  \begin{aligned}
     ||W^TW - I||_F = 
     \sqrt{tr\bigg((W^TW - I) (W^TW - I)^T\bigg)} = \\
     \sqrt{tr((W^TW)^2)-2tr(W^TW)+I} = \\
     \sqrt{\sum_{i=1}^{d_{out}} (\lambda^2_i(W^TW) - 2\lambda_i(W^TW) + 1)} = \\
     \sqrt{\sum_{i=1}^{d_{out}} (\lambda_i(W^TW) - 1)^2} 
  \end{aligned}
\end{equation}

To minimize the regularization term, the term $( \sum_{i=1}^{d_{out}} \lambda_i(W^TW)- 1)$ must become zero for all $i$'s which implies all $\lambda_i(W^T)$'s must equal 1. 

\begin{table*}[!h]
\centering
\small
\caption{Hyperparameter setting used for training.}
\begin{tabular}{c|ccccccc} 
\hline
Dataset                   & $\mu_{base}$ & \begin{tabular}[c]{@{}c@{}}Embedding \\ Size\end{tabular} & \begin{tabular}[c]{@{}c@{}}Encoder Hidden\\ Sizes\end{tabular} &  \begin{tabular}[c]{@{}c@{}}Predictor Hidden\\ Sizes\end{tabular}  & Batch Norm & Layer Norm   \\ 
\hline
\textbf{Cora}             & $10^{-4}$    & 128            & 256                  & 128                    & N          & N     \\
\textbf{CiteSeer}         & $10^{-3}$    & 256            & 512                  & 256                    & N          & N     \\
\textbf{PubMed}           & $10^{-4}$    & 256            & 512                  & 256                    & N          & N    \\
\textbf{DBLP}             & $10^{-3}$    & 256            & 512                  & 256                    & N          & N     \\
\textbf{Amz Computers} & $5.10^{-4}$  & 128            & 256                  & 512                    & Y          & N        \\
\textbf{Amz Photos}    & $10^{-4}$    & 256           & 512                 & 512                    & Y          & N        \\
\textbf{Amz Ratings}   & $10^{-4}$    & 256            & 512                  & 512                    & N          & N         \\
\textbf{CoCS}      & $10^{-5}$    & 256            & 512                  & 512                    & Y          & N       \\
\textbf{CoPhysics} & $10^{-5}$    & 128            & 256                  & 512                   & Y          & N          \\
\textbf{WikiCS}           & $5. 10^{-4}$ & 256            & 512                  & 512                    & Y          & N                        \\
\textbf{RomanEmpire}      & $10^{-4}$    & 128            & 256                  & 128                    & N          & N                         \\
\textbf{PPI}              & $5.10^{-3}$  & 512            & 512,512              & 512                    & N          & Y                        \\
\textbf{ogbn-arXiv}       & $10^{-2}$    & 256            & 256,256              & 256                    & Y         & N                 
\end{tabular}
\label{ImplementationTable}
\end{table*}

\subsection{WERank Encourages Short Mappings}
In this section, we shed some light on how WERank impacts the geometry of the representation space. We turn our attention to the Lipschitz constant of a layer $W$, which is the smallest upper bound on the amount of change in the output of the layer after changing the input. For any two inputs $h_i, h_j$, we can write the Lipschitz constant as the smallest $L$ such that 
\begin{equation*}
     \frac{\| (h_i - h_j) W \|}{\| h_i - h_j \|} \leq L
\end{equation*}

We can  show that in our case, the  Lipschitz constant is the largest singular value of $W$:
\begin{equation}
    \begin{aligned}
         \frac{\| (h_i - h_j) W \|}{\| h_i - h_j \|} \leq
         \sup_{h_i \neq h_j}  \frac{\| (h_i - h_j) W \|}{\| h_i - h_j \|} = 
         \max_{ \| h_i - h_j \| = 1} \|(h_i - h_j) W \| =
         \sigma_{max}(W) = \sigma_1        
    \end{aligned}
\end{equation}

The above holds since $\max_{ \|v\| = 1} \| vW \|$ is the norm definition of the largest singular value of $W$. Similarly, we can write:

\begin{equation*}
    \begin{aligned}
         \frac{\| (h_i - h_j) W \|}{\| h_i - h_j \|} \geq
         \inf_{h_i \neq h_j}  \frac{\| (h_i - h_j) W \|}{\| h_i - h_j \|} = 
         \min_{ \| h_i - h_j \| = 1} \|(h_i - h_j) W \| =
         \sigma_{min}(W) = \sigma_{d_{out}}        
    \end{aligned}
\end{equation*}

We note that WERank encourages $\sigma_1,...,\sigma_{d_{out}}$ to be close to 1. Thus, WERank encourages $W$ not to increase or decrease any distance (in other words, $W$ is encouraged to be a short map). 

It is not hard to see that the same argument can be made about a deep network $f$ consisting of parameters $W_1, ..., W_L$. Since the Lipschitz constant of the composition of functions is less than or equal the multiplication of Lipschitz constants of each function, we can write:

\begin{equation*}
    \begin{aligned}
         \frac{\| f(h_i) - f(h_j) \|}{\| h_i - h_j \|} \leq
         \sup_{h_i \neq h_j}  \frac{\| (h_i - h_j) W_1 \|}{\| h_i - h_j \|} \times ... \times \frac{\| (h_i - h_j) W_L \|}{\| h_i - h_j \|} \leq 
         \sigma_{max}(W_1) \times ... \times \sigma_{max}(W_L)        
    \end{aligned}
\end{equation*}

Since WERank pushes $\sigma_{max}(W_i)$ to get close to 1 for all $i= 1,...,L$, the product $\sigma_{max}(W_1) \times ... \times \sigma_{max}(W_L)$ is also pushed to be close to 1. A similar argument can be made to derive $\sigma_{min}(W_1) \times ... \times \sigma_{min}(W_L)$ as a lower bound on $\frac{\| f(h_i) - f(h_j) \|}{\| h_i - h_j \|}$. As a result, the network $f$ would be a distance preserving function.

In general, we can conclude that WERank results in the model responding more conservatively to changes in the inputs. We leave further investigation on the implications of WERank on the geometry of the representation space as future work. 

\subsection{Computational Complexity}
For a network with layers $W_1, ..., W_L$ with $W_l \in \mathbb{R}^{d_l \times d_{l+1}}$, the complexity of computing WERank (i.e. $\| WW^T - I \|$) at the $l_{th}$ layer is $O(d_l . d_{l+1} ^2)$ (through the naive matrix multiplication algorithm). Thus, the complexity of computing WERank on the entire network is $\sum_{l = 1}^{L} O(d_l . d^2_{l+1}) = \max_l O(d_l . d_{l+1}^2)$. We note that the cost of computing the covariance term in VICReg is $O(N.d_{out}^2)$, where $N$ is the number of data points ($N$ usually being much larger than the hidden feature dimensions). 

\section{Implementation Details}

For scientific accuracy, our proposed setup stays close to the setting described by BGRL \cite{thakoor2022largescale}. We refer the reader to the original BGRL paper for details on optimization settings and frozen linear evaluation of embeddings. Table \ref{ImplementationTable} describes the hyperparameter and architectural details of our implementation. We note that all encoders are GCNs, except PPI which uses MeanPooling encoders with residual connections. We use PReLU activation\cite{iccv_HeZRS15} in all experiments. We note that $\mu_{base}$ denotes the base learning rate for the AdamW optimizer \cite{DBLP:journals/corr/KingmaB14}.

Table \ref{augFactorTable} outlines the distortion paramaters used for each dataset. We refer the interested reader to Appendix D of the work by \cite{thakoor2022largescale} for detailed explanation on how node feature and edge masking are done.

\begin{table}[!h]
\centering
\small
\caption{Distortion parameters used for training. \\ 
}
\begin{tabular}{c|cccc}
\hline
Dataset          & $p_{f, 1}$ & $p_{f, 2}$ & $p_{e,1}$ & $p_{e,2}$ \\ \hline
Cora             & 0.2         & 0.3         & 0.4        & 0.4        \\
CiteSeer         & 0.2         & 0.3         & 0.0        & 0.2        \\
PubMed           & 0.4         & 0.0         & 0.1        & 0.2        \\
DBLP             & 0.1         & 0.1         & 0.4        & 0.0        \\
Amazon Computers & 0.2         & 0.1         & 0.5        & 0.4        \\
Amazon Photos    & 0.1         & 0.2         & 0.4        & 0.1        \\
Amazon Ratings   & 0.3         & 0.1         & 0.1        & 0.3        \\
Coauthor CS      & 0.3         & 0.4         & 0.4        & 0.1        \\
Coauthor Physics & 0.1         & 0.4         & 0.4        & 0.1        \\
WikiCS           & 0.2         & 0.1         & 0.2        & 0.3        \\
RomanEmpire      & 0.3         & .01         & 0.2        & 0.2        \\
PPI              & 0.25        & 0.00        & 0.30       & 0.25       \\
ogbn-arxiv       & 0.0         & 0.0         & 0.6        & 0.6       
\end{tabular}
\label{augFactorTable}
\end{table}

\section{Dataset Details}
\label{App:Datasets}

\begin{table*}[!t]
\centering
\small
\caption{Statistics of datasets used in the experiments. \\ 
}
\begin{tabular}{cccccccc}
\hline
Size                                        & Dataset          & Type         & Task         & Nodes   & Edges     & Features & Classes          \\ \hline
\multicolumn{1}{c|}{\multirow{5}{*}{Small}} & \textbf{Cora}             & Homophilic   & Transductive & 2,708   & 5,421     & 1,433    & 7                \\
\multicolumn{1}{c|}{}                       & \textbf{CiteSeer}         & Homophilic   & Transductive & 3,327   & 4,732     & 3,703    & 6                \\
\multicolumn{1}{c|}{}                       & \textbf{PubMed}           & Homophilic   & Transductive & 19,717  & 44,338    & 500      & 3                \\
\multicolumn{1}{c|}{}                       & \textbf{DBLP}             & Homophilic   & Transductive & 17,716  & 105,734   & 1,639    & 4                \\ \hline
\multicolumn{1}{c|}{}                 & \textbf{Amazon Computers} & Homophilic   & Transductive & 13,752  & 245,861   & 767      & 10               \\
\multicolumn{1}{c|}{}                       & \textbf{Amazon Photos}    & Homophilic   & Transductive & 7,650   & 119,081   & 745      & 8                \\
\multicolumn{1}{c|}{}                       & \textbf{Amazon Ratings}    & Heterophilic   & Transductive & 24492  &  93050  & 300  &    5             \\
\multicolumn{1}{c|}{Medium}                 & \textbf{Coauthor CS}      & Homophilic   & Transductive & 18,333  & 81,894    & 6,805    & 15               \\
\multicolumn{1}{c|}{}                       & \textbf{Coauthor Physics} & Homophilic   & Transductive & 34,493  & 247,962   & 8,415    & 5                \\
\multicolumn{1}{c|}{}                       & \textbf{WikiCS}           & Homophilic   & Transductive & 11,701  & 216,213   & 300      & 10               \\
\multicolumn{1}{c|}{}                       & \textbf{RomanEmpire}      & Heterophilic & Transductive & 22,662  & 32,927    & 300      & 18               \\
\hline
\multicolumn{1}{c|}{Large}                       & \textbf{PPI}              & Homophilic   & Inductive    & 56,944  & 818,716   & 50       & 121 (multilabel) \\ 
\multicolumn{1}{c|}{}             & \textbf{ogbn-arxiv}       & Homophilic   & Transductive & 169,343 & 1,166,243 & 128      & 40        
\end{tabular}
\label{stats_datasets}
\end{table*}

We provide a brief overview of the benchmark datasets as shown in Table \ref{stats_datasets}. Unless otherwise specified, we follow the work by \cite{thakoor2022largescale} and randomly split the nodes into (10/10/80\%) train/validation/test sets.

\textbf{Cora, Citeseer, Pubmed,and DBLP: } Graphs are constructed from computer science article citation links, where nodes correspond to articles and edges correspond to citation links between articles. The task is to predict predicting article subject categories. Each node has a sparse bag-of-words as feature \cite{SenRad_2008,bojchevski2018deep}.

\textbf{WikiCS: } Nodes represent Wikipedia articles about Computer Science and edges representing links between them. The task is to predict the article sub-field. Node features are the average of GloVE \cite{DBLP:conf/emnlp/PenningtonSM14} embeddings of all words in the article. WikiCS comes with 20 canonical train/valid/test splits, which we directly use. \footnote{https://github.com/pmernyei/wiki-cs-dataset/tree/master/dataset}

\textbf{Amazon Computers, Amazon Photos: } Nodes represents products in the Amazon co-purchase graph \cite{DBLP:conf/sigir/McAuleyTSH15}. An edge exists between pairs of goods frequently purchased together. The task is to predict the product category. Node features are a bag-of-words representation of a product’s reviews. \footnote{https://github.com/shchur/gnn-benchmark/tree/master/data/npz}

\textbf{Amazon Ratings: }Introduced by \cite{platonov2023a}, nodes are products such as DVDs, books, and VHS video tapes on the Amazon product co-purchasing network metadata dataset. Two nodes have an edge if the corresponding products are frequently bought together. Node features are the mean of FastText embeddings \cite{grave-etal-2018-learning} for words in the product description. The task is to predict the average rating given to a product by reviewers.

\textbf{Coauthor CS, Coauthor Physics: } Nodes represent authors in the Microsoft Academic Graph \cite{10.1145/2740908.2742839}, and edges are drawn between authors who have co-authored a paper. Node features are a bag-of-words representation of the keywords of an author’s papers. The task is to predict the author research field. 

\textbf{Roman Empire: } Each node in the graph corresponds to one (non-unique) word in the Roman Empire article from English Wikipedia. Two words are connected with an edge if either they follow each other in the text, or they are connected in the dependency tree of the sentence. The class of a node is its syntactic role. Full dataset details can be found in the paper by \cite{platonov2023a}. 

\textbf{ogbn-arXiv: } Nodes represent computer science papers on arXiv indexed by the Microsoft Academic Graph \cite{10.1145/2740908.2742839}. An edge is drawn between two nodes if one node has cited the other. Node features consists of the average word-embedding of all words in the paper, where the embeddings are computed using a skip-gram model \cite{DBLP:journals/corr/abs-1301-3781} over the entire text corpus. The task is to predict the paper arXiv subject area.

\textbf{PPI: } This dataset consists of 24 graphs of protein-protein interaction networks \cite{10.1093/bioinformatics/btx252, he2015delving}, each corresponding to different human tissues. Node features are computed from various biological properties. The task is multilabel classification, where each node can have up to 121 labels. Similar to \cite{thakoor2022largescale}, we use the standard dataset split as 20 graphs for training, 2 for validation, and 2 for testing. 


\subsection{Graph Related Definitions}
\textbf{Transductive vs. Inductive: }For transductive learning tasks, features of all data are used, but the labels of the test set are masked during training. For inductive learning tasks, tests are conducted on unseen or untrained nodes and graphs.

\textbf{Homophilic vs. Heterophilic: } A homophilic graph is one where nodes are connected with semantically similar nodes. Thus, most regions of a homophilic graph would predominantly share the same label. Heterophilic graph do not share this trait; nodes sharing the same label would be distributed in different regions of the graph.

\section{Toy Dataset Ablations}
\label{aapA}

\begin{figure*}[!h]
  \centering
    \includegraphics[width=0.25\textwidth]{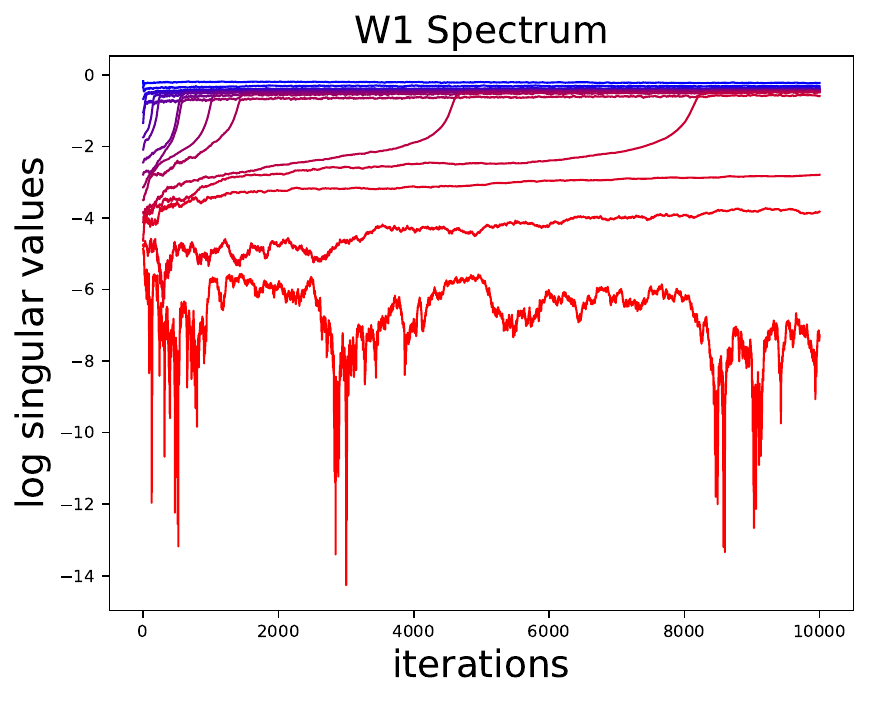}
    \includegraphics[width=0.25\textwidth]{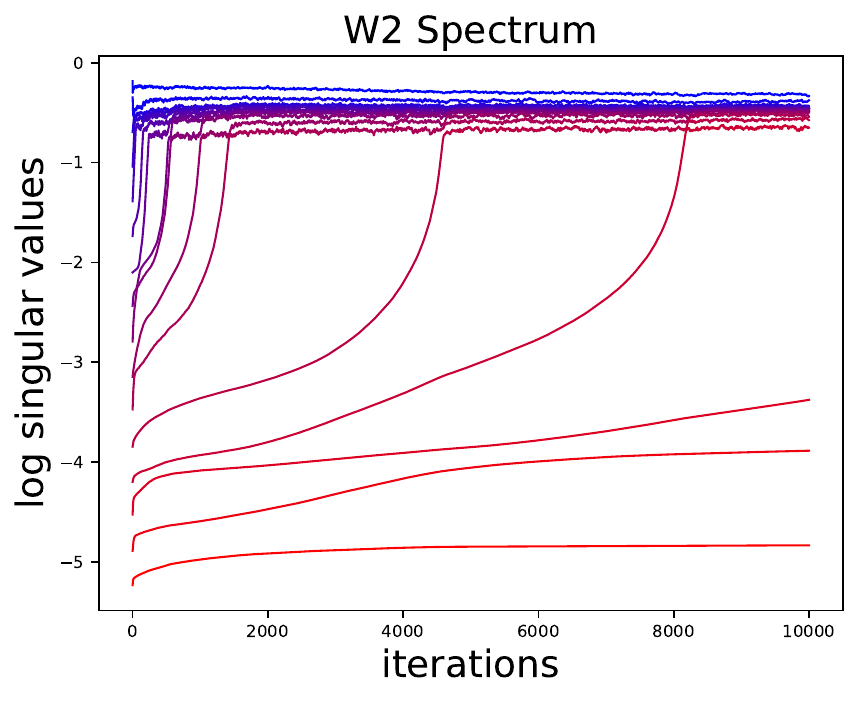}
    \includegraphics[width=0.25\textwidth]{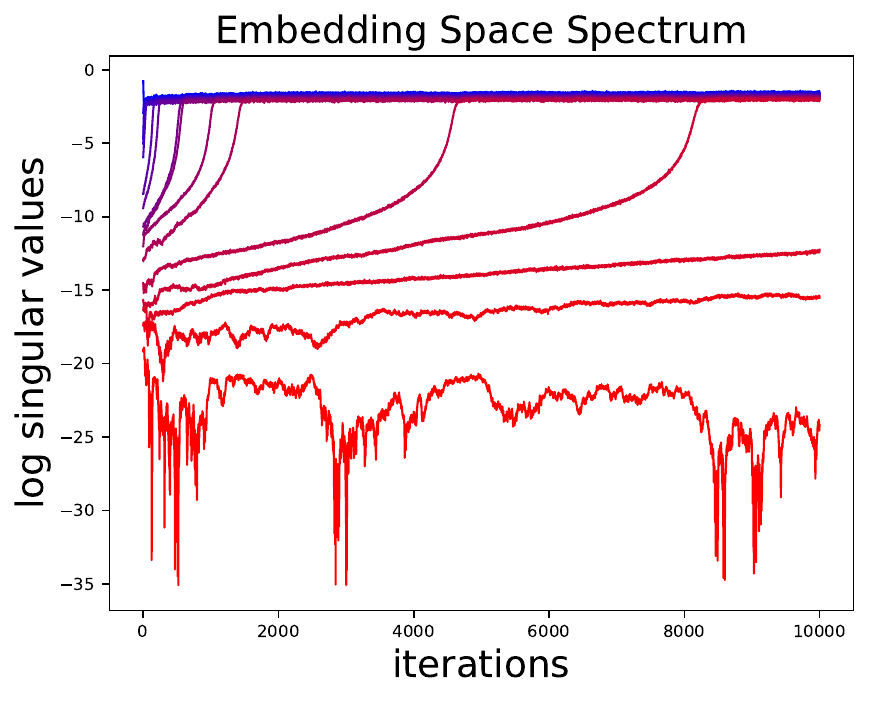}
    \includegraphics[width=0.25\textwidth]{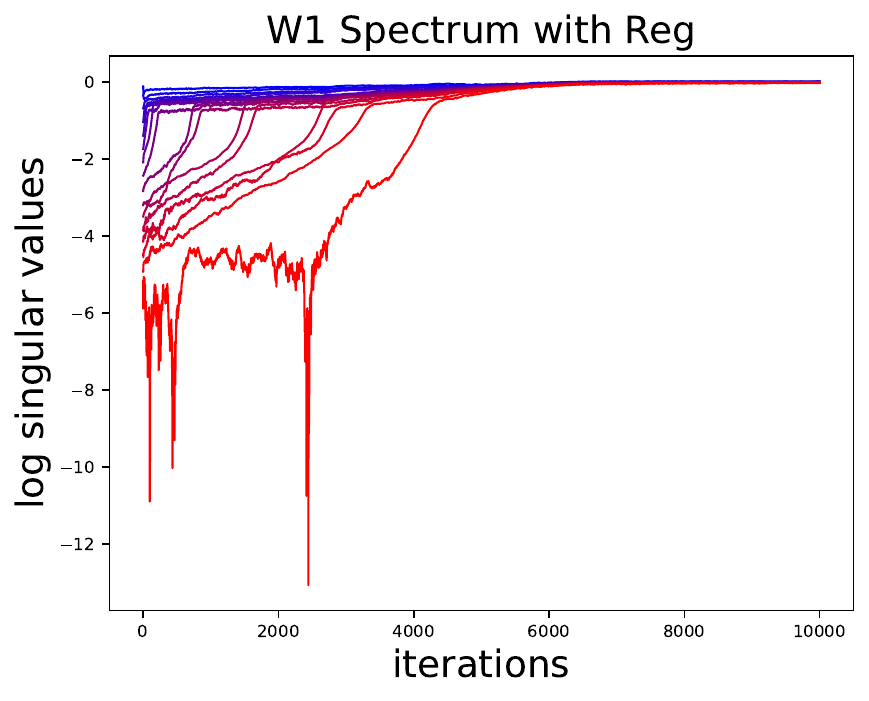}
    \includegraphics[width=0.25\textwidth]{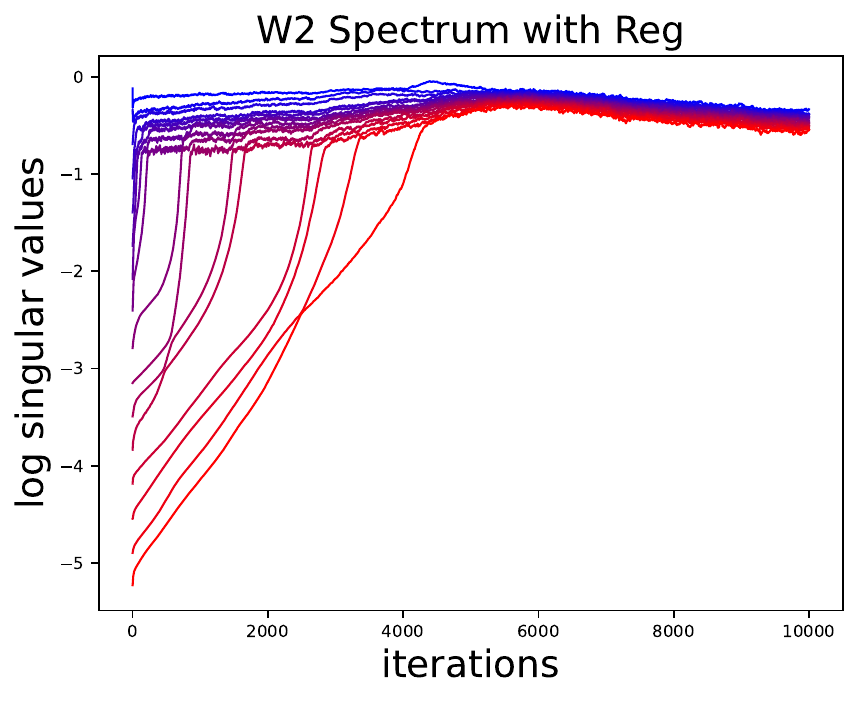}
    \includegraphics[width=0.25\textwidth]{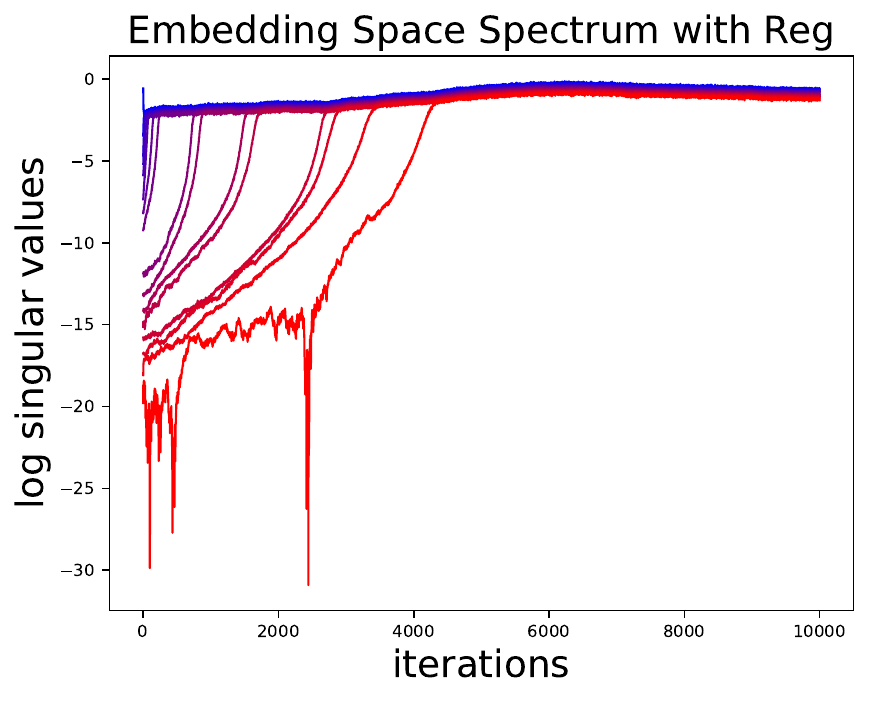}
  \caption{The singular values of the weight matrices and the embedding space covariance matrix during training (top) InfoNCE model with no regularization (button) InfoNCE model with the WERank regularizer. The augmentation magnitude ($k$) is set to $0.1$}
\label{extra-toy-info}
\end{figure*}

\begin{figure*}[!h]
  \centering
    \includegraphics[width=0.25\textwidth]{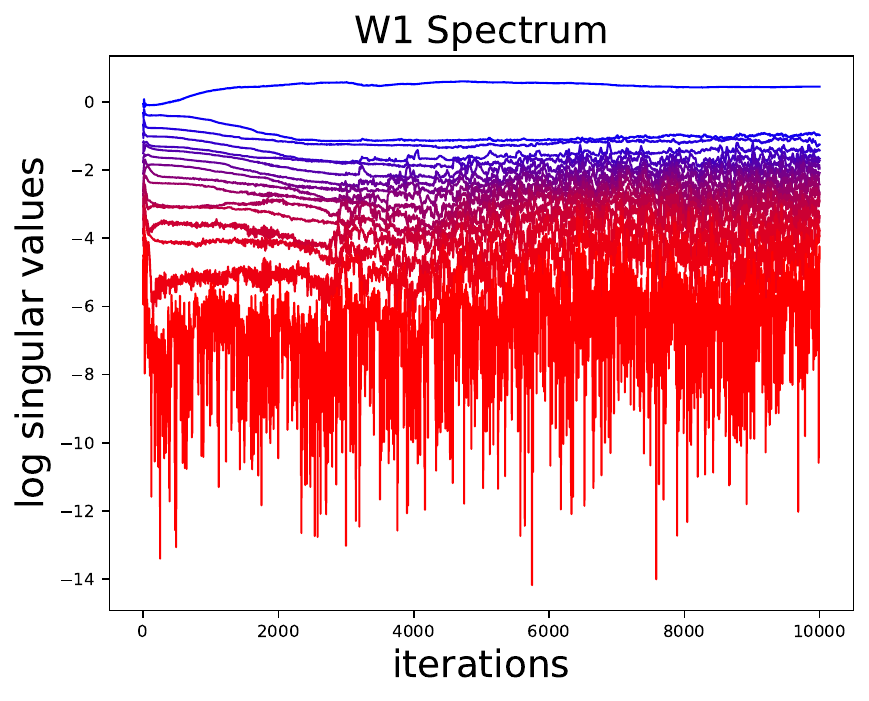}
    \includegraphics[width=0.25\textwidth]{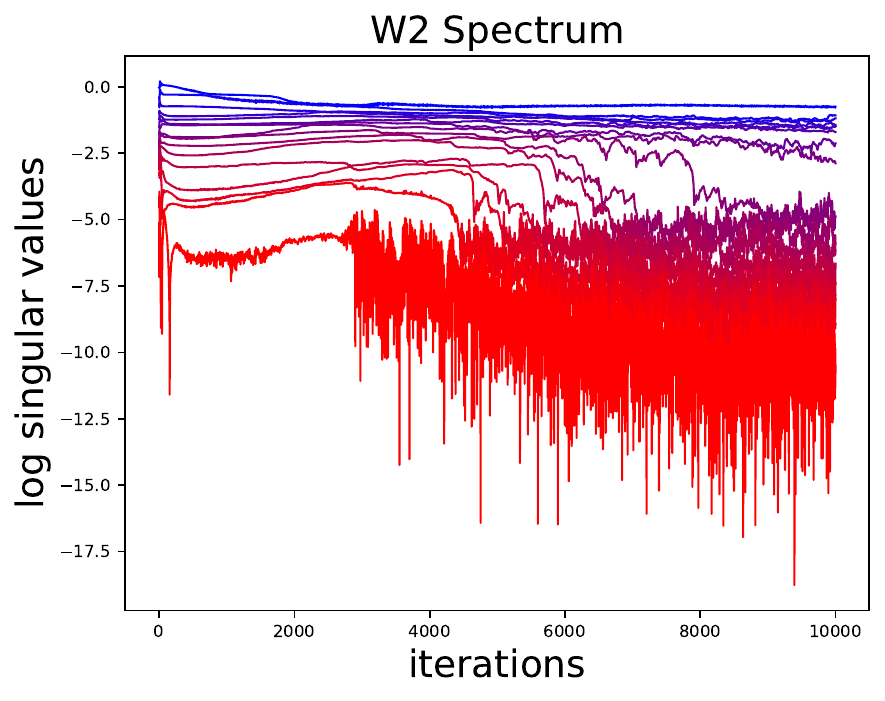}
    \includegraphics[width=0.25\textwidth]{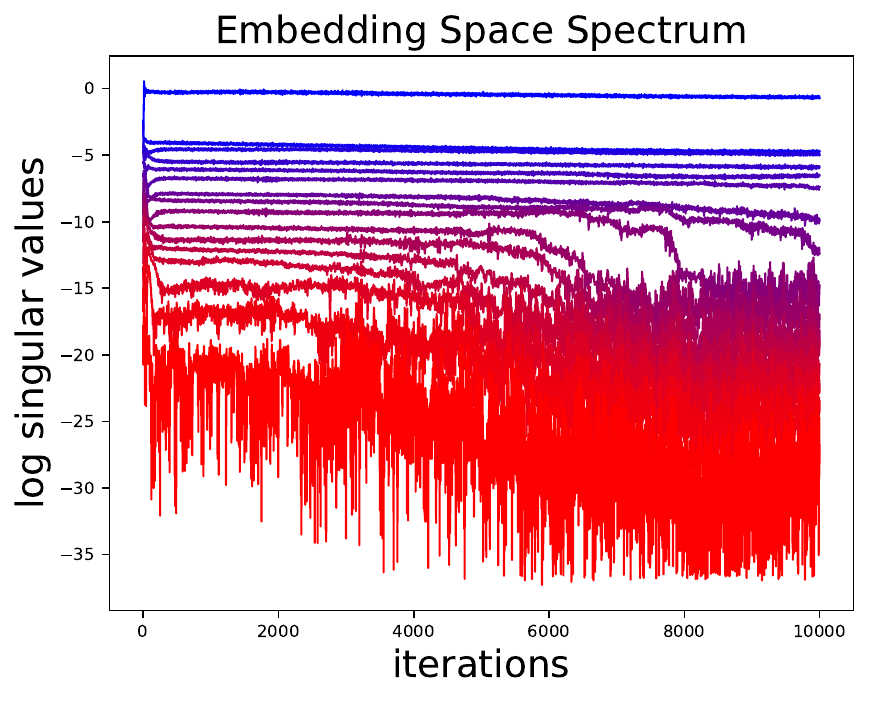}
    \includegraphics[width=0.25\textwidth]{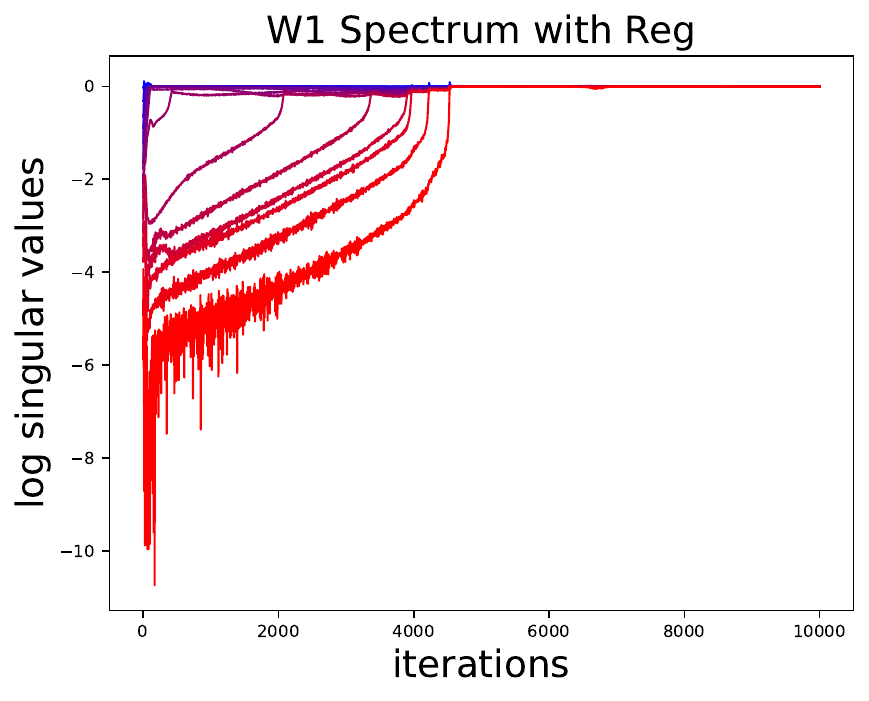}
    \includegraphics[width=0.25\textwidth]{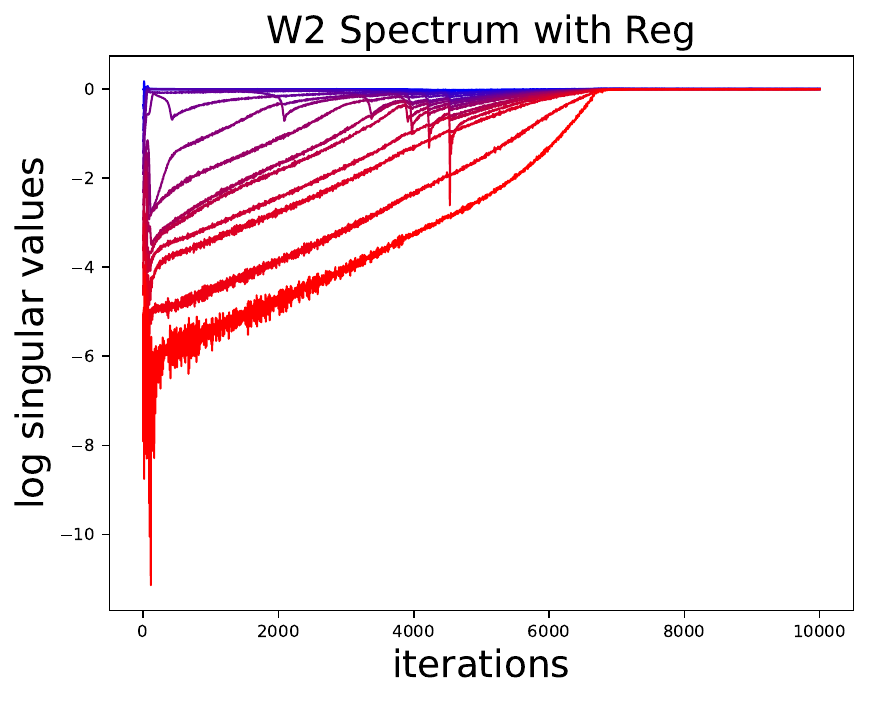}
    \includegraphics[width=0.25\textwidth]{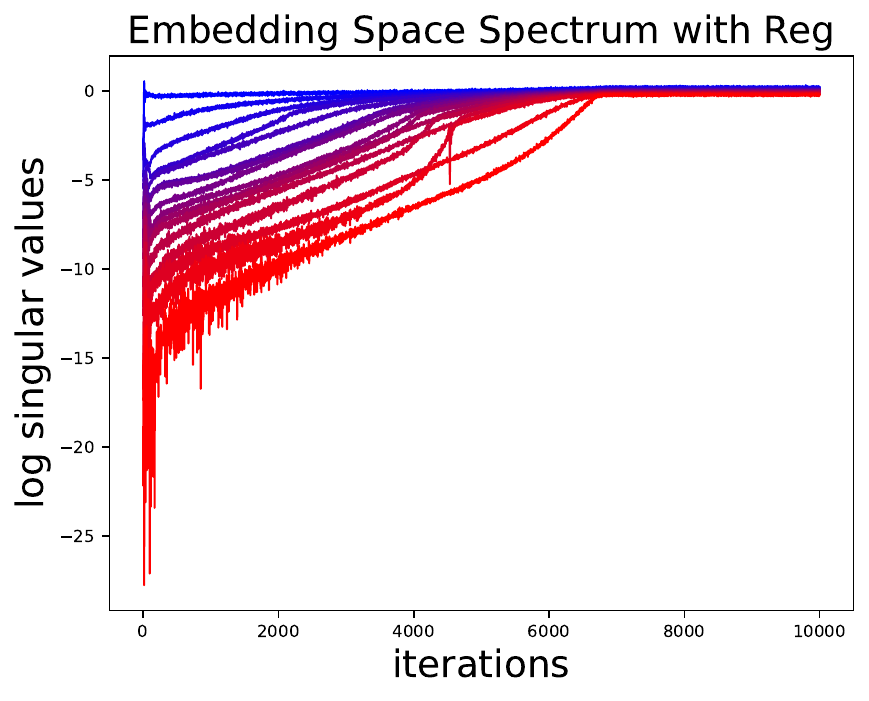}
  \caption{The singular values of the weight matrices and the embedding space covariance matrix during training (top) EMA model with no regularization (button) EMA model with the WERank regularizer. The augmentation magnitude ($k$) is set to $0.1$}
\label{extra-toy-ema}
\end{figure*}

\subsection{Singular Value Specturum of InfoNCE and EMA Models}
In addition to the VICReg model, we provide plots depicting the evolution of singular values in InfoNCE and EMA models in Figure \ref{extra-toy-info} and \ref{extra-toy-ema} respectively. All models are trained for 10000 epochs in the full batch regime. The variance, invariance, covariance coefficients for VICReg are set to 10, 10 and 1 respectively. The EMA model is implemented in a similar fashion to BYOL  \cite{DBLP:journals/corr/abs-2006-07733}, with a decay rate of 0.995.

We apply basic stochastic gradient descent without momentum or weight decay when training the VICReg and InfoNCE models. However, we find that applying the same optimizer on the EMA model results in the singular values remaining constant throughout training. Thus, we apply the AdamW \cite{DBLP:journals/corr/KingmaB14} optimizer with a learning rate of 0.01 and weight decay 0.0003. \footnote{In general, the traning dynamics of the EMA model is highly sensitive to the choice of the optimizer. However, changing the hyperparameters of the AdamW optimizer does not result in the singular values converging to 1.}

The WERank coefficient is set to 0.1 for VICReg and InfoNCE models. We find that WERank has an immediate impact on the EMA model, where it pushes all the singular values to one within a few hundred epochs. As such, we set the WERank coefficient to 0.02 for the EMA model (for illustrative purposes). 

We notice that the first eight singular values converge to one in VICReg and InfoNCE models. However, the EMA model has trouble pushing the first eight singular values higher. We suspect this is due to the lack of a mechanism to encourage high rank representations in the EMA model.

\begin{figure*}[!h]
  \centering
   \includegraphics[width=0.32\textwidth]
    {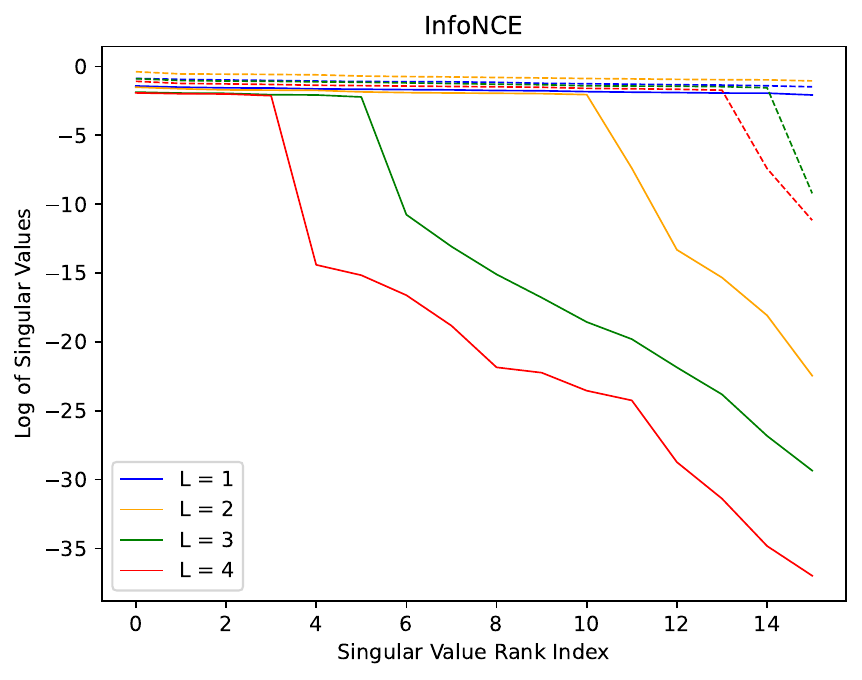}
    \includegraphics[width=0.32\textwidth]{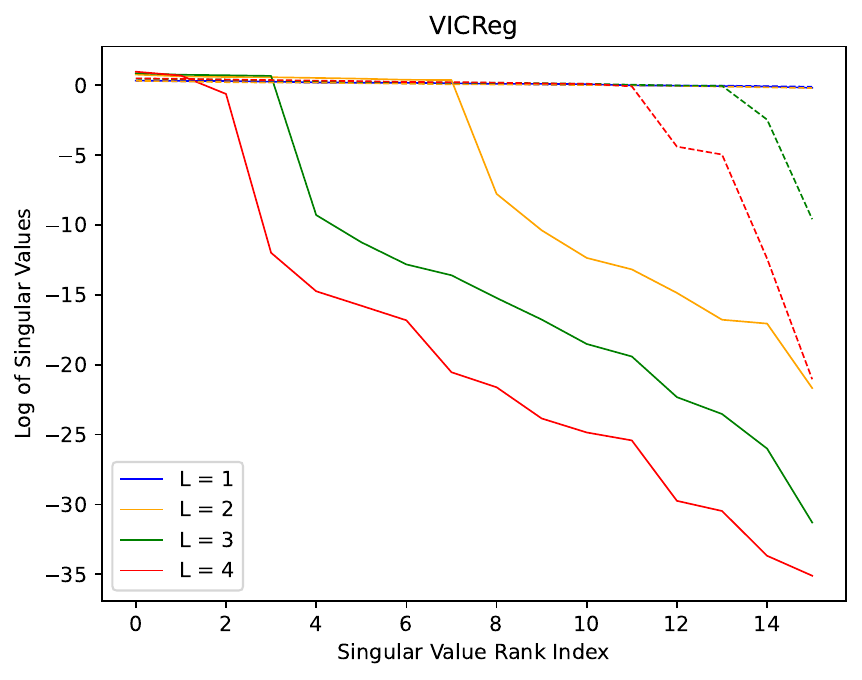}
    \includegraphics[width=0.32\textwidth]{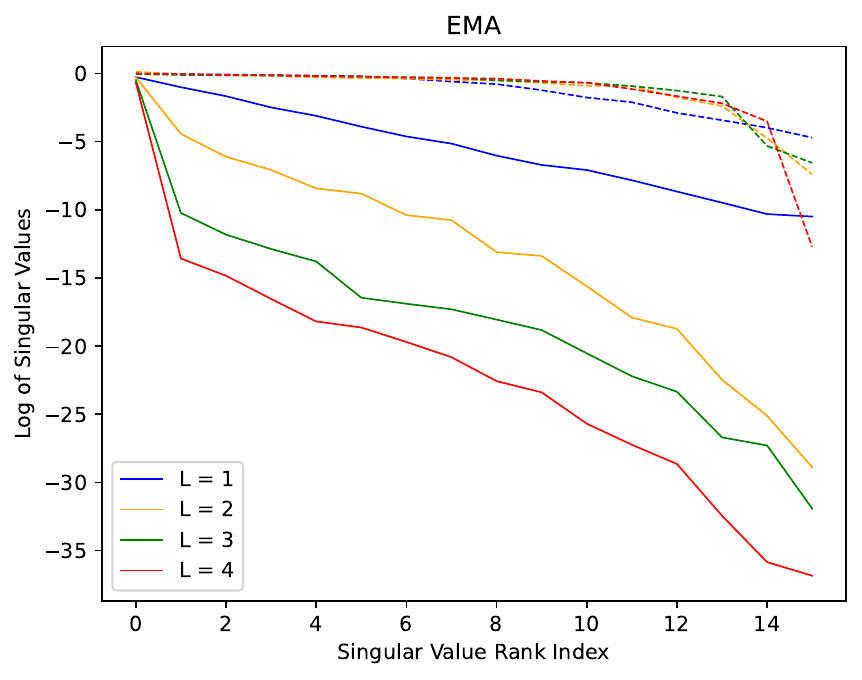}
  \caption{Embedding space singular values after training the models for 5000 epochs. L denotes the number of layers, dashed lines depict the model with WERank regularization. Training each model + WERank for 10000 epochs would result in all singular values converging to 1, while models without WERank face rank degradation.}
  \label{depthFig}
\end{figure*}

\begin{table*}[!h]
  \centering
  \small
  \caption{Downstream Performance of BGRL and BGRL + WERank under different magnitudes of augmentation. The experiments are over 20 random dataset splits and model initializations. \\}
\begin{tabular}{l|ccccccc} 
\hline
Dataset                           & Model         & 0.01 & 0.05 & 0.1 & 0.5 & 2  \\ 
\hline
\multirow{2}{*}{Amz Ratings}   & BGRL          &   $42.60 \pm 0.02$   &  $42.62 \pm 0.02$    &  $42.51 \pm 0.02$   &  $43.09 \pm 0.01$   &   $\mathbf{43.68 \pm 0.01}$ \\
                                & BGRL + WERank &  $\mathbf{42.96 \pm 0.01}$    &  $\mathbf{42.94 \pm 0.01}$    &   $\mathbf{43.35 \pm 0.01}$  & $\mathbf{43.21 \pm 0.01}$ &   $43.57 \pm 0.01$  \\ 
\hline
\multirow{2}{*}{Roman Empire}     & BGRL          &  $65.74 \pm 0.21$  &  $66.37 \pm 0.17$    &  $67.95 \pm 0.21$   &  $73.58 \pm 0.2$   &   $\mathbf{70.68 \pm 0.01}$  \\
                                  & BGRL + WERank &    $\mathbf{66.18 \pm 0.01}$ &    $\mathbf{67.08 \pm 0.01}$  &   $\mathbf{68.41 \pm 0.03}$ &   $\mathbf{73.76 \pm 0.02}$  &   $70.43 \pm 0.01$  \\ 
\hline
\end{tabular}
\label{augTable2}
\end{table*}

\subsection{Rank Degradation Prevention in Deeper Models}

In Figure \ref{depthFig}, we plot the singular values of the embedding spaces of different models with and without the WERank regularization applied. Evidently, WERank is effective in preventing rank degradation in deeper networks, where the impact caused by implicit regularization aggravates.

\section{Additional Results}
\label{addResults}

\subsection{Results on Heterophilic Graphs}
In addition to the graphs used in the original BGRL paper and other SoTA graph SSL methods \cite{thakoor2022largescale,zhu2020deep,veličković2018deep}, we consider two heterophilic graphs for the following tasks: $(i)$ classifying product ratings on Amazon product co-purchasing network $(ii)$ Predicting synthetic roles of words in graph induced from the Roman Empire article from English Wikipedia. Detailed explanations on the datasets can be found in Appendix \ref{App:Datasets}. 

We include the data augmentation ablation and the downstream accuracy results in Table \ref{augTable2} and \ref{dwnTable4} respectively. We note that the lack of homophily in these graphs makes the prediction tasks more difficult compared to homophilic graphs. While WERank helps BGRL achieve higher accuracy on Amazon Ratings, the performance of BGRL with WERank falls lightly behind vanilla BGRL on the Roman Empire dataset. Our results on the augmentation ablation show that WERank is most helpful in the low augmentation regime.

\begin{table}[!h]
    \small
    \centering
    \caption{Classification accuracy along with standard deviations on the  Hetrophilic dataset. Our experiments are over 20 random dataset splits and model initializations. \\
    }
    \begin{tabular}{lcccccccc}
    \hline
        \textbf{Method/Dataset} &\textbf{Roman Empire} & \textbf{Amz Ratings}  \\ \hline
        \textbf{RandInit-MLP}  & 63.37 $\pm$ 0.01 & 39.00 $\pm$ 0.30 \\ 
        \textbf{Supervised} & 77.09 $\pm$ 0.04 & 43.10 $\pm$ 1.17  \\ 
        \hline
        \textbf{VICReg} & 58.13  $\pm$ 0.01 & 43.14  $\pm$ 0.01 \\ 
        \textbf{GBT} &  61.56 $\pm$ 0.01 & 42.68  $\pm$ 0.01   \\ 
        \textbf{GSwav} & 62.11 $\pm$ 0.00 & 42.41  $\pm$ 0.02 \\ 
        \textbf{Grace} & 63.28 $\pm$ 0.02 &  OOM \\ 
        \hline
        \textbf{BGRL} &  74.61 $\pm$ 0.01 & 42.86  $\pm$ 0.02\\ 
        \textbf{BGRL+WERank} & 74.44 $\pm$ 0.00  & 43.28  $\pm$ 0.00 \\ \hline
    \end{tabular}
    \label{dwnTable4}
\end{table}

\begin{figure*}[!h]
  \centering
    \includegraphics[width=0.20\textwidth]{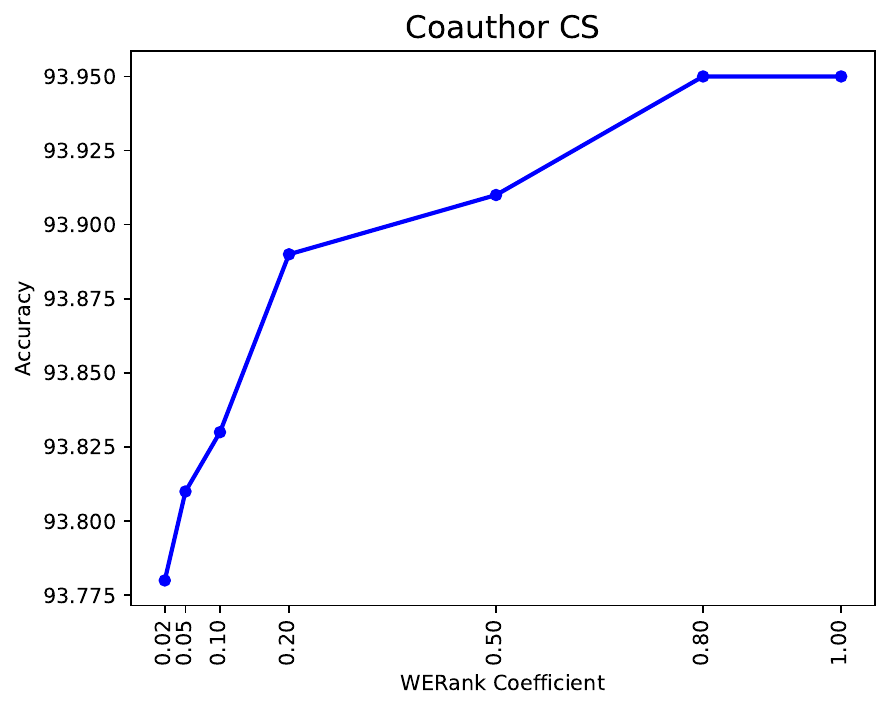}
    \includegraphics[width=0.20\textwidth]{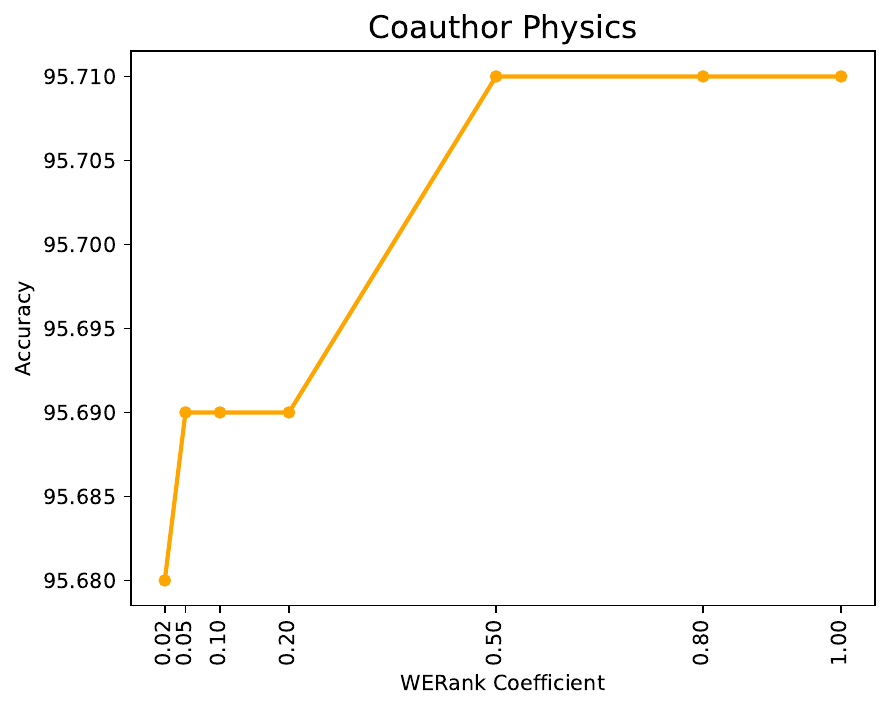}
    \includegraphics[width=0.20\textwidth]{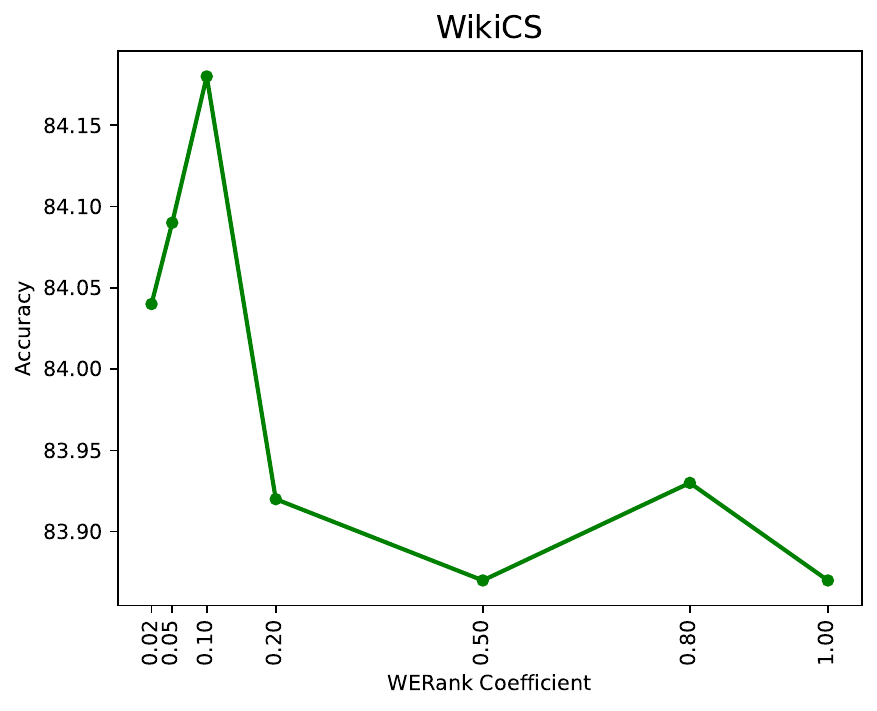}
    \includegraphics[width=0.20\textwidth]{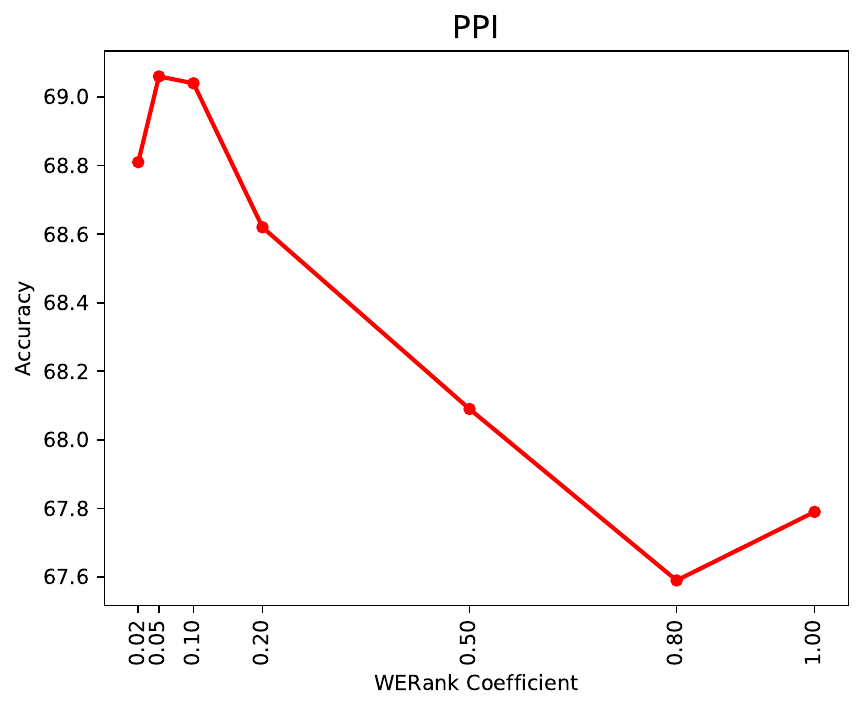}
  \caption{Downstream Accuracy of BGRL trained with different WERank coefficients.}
  \label{coeffFig}
\end{figure*}

\subsection{Results on Homogeneous Graphs}
We train BGRL with different coefficients for WERank. The same coefficient is applied to every layer of the encoder network. We include the results in Table \ref{tableCoeff} with visualizations in Figure \ref{coeffFig}.

\begin{table*}[!h]
\small
\centering
\caption{BGRL with WERank trained with different regularization coefficients $\alpha$. The same coefficient $\alpha$ is applied to every encoder layer. The experiments are over 20 random dataset splits and model initializations. \\}
\begin{tabular}{c|ccccc} 
\hline
Layer coefficient & Coauthor CS      & Coauthor Physics & WikiCS           & PPI              & ogbn Arxiv        \\ 
\hline
0.02              & $93.78 \pm 0.37$ & $95.68 \pm 0.14$ & $84.05 \pm 0.17$ & $68.81 \pm 0.01$ & $71.53 \pm 0.03$  \\
0.05              & $93.81 \pm 0.35$ & $95.69 \pm 0.13$ & $84.10 \pm 0.18$ & $69.07 \pm 0.02$ & $71.62 \pm 0.05$  \\
0.1               & $93.83 \pm 0.34$ & $95.69 \pm 0.14$ & $84.18 \pm 0.22$ & $69.04 \pm 0.01$ & $71.44 \pm 0.03$  \\
0.2               & $93.89 \pm 0.34$ & $95.69 \pm 0.14$ & $83.90 \pm 0.16$ & $68.62 \pm 0.01$ & $71.37 \pm 0.03$  \\
0.5               & $93.91 \pm 0.39$ & $95.71 \pm 0.14$ & $83.87 \pm 0.24$ & $68.09 \pm 0.01$ & $71.34 \pm 0.03$  \\
0.8               & $93.95 \pm 0.42$ & $95.71 \pm 0.14$ & $83.93 \pm 0.19$ & $67.59 \pm 0.01$ & $71.25 \pm 0.03$  \\
1                 & $93.95 \pm 0.42$ & $95.71 \pm 0.14$ & $83.87 \pm 0.19$ & $67.79 \pm 0.01$ & $71.27 \pm 0.03$ 
\end{tabular}
\label{tableCoeff}
\end{table*}

As illustrated in Table \ref{variable_coeffFig}, to better understand the impact of different WERank coefficients per layer, we consider a 3-layer GCN model with the five benchmark datasets. We train two models per dataset; in one model we apply WERank coefficients 0.2, 0.1, 0.05 to the first, second, and third layers and in another we apply them in reverse. Our results hint that applying a higher regularization factor the final layers is more impactful than applying higher factors in early layers. In datasets where higher regularization results in higher accuracy (CoPhy and CoCS), applying higher factors in early layers results in slightly higher accuracy. On the other hand, when the dataset does not benefit from strong WERank regularization (ogbn-arXiv, PPI), applying higher factors in early layers results in lower accuracy.

\begin{table}[!h]
\centering
\caption{BGRL with WERank trained with different regularization coefficients $\alpha$ at each layer. The experiments are over 5 random dataset splits and model initializations.\\}
\begin{tabular}{lcc}
\hline
Dataset        & 0.2, 0.1, 0.05 & 0.05, 0.1, 0.2  \\ \hline
WikiCS         & 82.88 $\pm$ 0.10 & 82.90 $\pm$ 0.09 \\
CoPhy         &  95.44 $\pm$ 0.11 & 95.48 $\pm$ 0.10\\
CoCS          & 92.39 $\pm$ 0.26 & 92.61 $\pm$ 0.35 \\
PPI          &  69.12 $\pm$  0.04 & 69.03 $\pm$ 0.05 \\
ogbn-arXiv     & 71.65 $\pm$ 0.03 & 71.40 $\pm$ 0.04 \\
\end{tabular}
\label{variable_coeffFig}
\end{table}

\vfill

\clearpage

\section*{Checklist}



 \begin{enumerate}

 \item For all models and algorithms presented, check if you include:
 \begin{enumerate}
   \item A clear description of the mathematical setting, assumptions, algorithm, and/or model. [\textbf{Yes}/No/Not Applicable]
   \item An analysis of the properties and complexity (time, space, sample size) of any algorithm. [\textbf{Yes}/No/Not Applicable]
   \item (Optional) Anonymized source code, with specification of all dependencies, including external libraries. [Yes/No/Not Applicable]
 \end{enumerate}

 \item For any theoretical claim, check if you include:
 \begin{enumerate}
   \item Statements of the full set of assumptions of all theoretical results. [\textbf{Yes}/No/Not Applicable]
   \item Complete proofs of all theoretical results. [\textbf{Yes}/No/Not Applicable]
   \item Clear explanations of any assumptions. [\textbf{Yes}/No/Not Applicable]     
 \end{enumerate}

 \item For all figures and tables that present empirical results, check if you include:
 \begin{enumerate}
   \item The code, data, and instructions needed to reproduce the main experimental results (either in the supplemental material or as a URL). [\textbf{Yes}/No/Not Applicable]
   \item All the training details (e.g., data splits, hyperparameters, how they were chosen). [\textbf{Yes}/No/Not Applicable]
         \item A clear definition of the specific measure or statistics and error bars (e.g., with respect to the random seed after running experiments multiple times). [\textbf{Yes}/No/Not Applicable]
         \item A description of the computing infrastructure used. (e.g., type of GPUs, internal cluster, or cloud provider). [\textbf{Yes}/No/Not Applicable]
 \end{enumerate}

 \item If you are using existing assets (e.g., code, data, models) or curating/releasing new assets, check if you include:
 \begin{enumerate}
   \item Citations of the creator If your work uses existing assets. [\textbf{Yes}/No/Not Applicable]
   \item The license information of the assets, if applicable. [Yes/No/\textbf{Not Applicable}]
   \item New assets either in the supplemental material or as a URL, if applicable. [Yes/No/\textbf{Not Applicable}]
   \item Information about consent from data providers/curators. [Yes/No/\textbf{Not Applicable}]
   \item Discussion of sensible content if applicable, e.g., personally identifiable information or offensive content. [Yes/No/\textbf{Not Applicable}]
 \end{enumerate}

 \item If you used crowdsourcing or conducted research with human subjects, check if you include:
 \begin{enumerate}
   \item The full text of instructions given to participants and screenshots. [Yes/No/\textbf{Not Applicable}]
   \item Descriptions of potential participant risks, with links to Institutional Review Board (IRB) approvals if applicable. [Yes/No/\textbf{Not Applicable}]
   \item The estimated hourly wage paid to participants and the total amount spent on participant compensation. [Yes/No/\textbf{Not Applicable}]
 \end{enumerate}

 \end{enumerate}

\end{document}